\documentclass{article}

\PassOptionsToPackage{numbers, compress}{natbib}

\usepackage[preprint]{neurips_2026}

\usepackage[utf8]{inputenc} 
\usepackage[T1]{fontenc}    
\usepackage{hyperref}       
\usepackage{url}            
\usepackage{booktabs}       
\usepackage{amsfonts}       
\usepackage{nicefrac}       
\usepackage{microtype}      
\usepackage{xcolor}         
\usepackage{amsmath}
\usepackage{amssymb}
\usepackage{dsfont}
\usepackage{algorithm}
\usepackage{algpseudocode}
\usepackage{subcaption}
\usepackage{xspace}

\usepackage{tikz}
\usepackage{multirow} 
\usepackage{bigdelim} 

\definecolor{GalacticPurple}{HTML}{6A00F4}
\definecolor{NeonMagenta}{HTML}{FF007F}
\definecolor{NebulaPink}{HTML}{FF7EEB}
\definecolor{StellarCyan}{HTML}{00F0FF}
\definecolor{SupernovaGold}{HTML}{FFB300}
\definecolor{PlasmaBlue}{HTML}{0B1D51}
\definecolor{SubBlue}{HTML}{6699ff}
\definecolor{InsGreen}{HTML}{99cc99}
\definecolor{DelRed}{HTML}{993333}

\newcommand{\ins}{\textcolor{InsGreen}{\text{ins}}}
\newcommand{\del}{\textcolor{DelRed}{\text{del}}}
\newcommand{\sub}{\textcolor{SubBlue}{\text{sub}}}
\newcommand{\morph}{\textbf{Morph}\xspace}

\title{Generative Molecular Morphing for Flexible-Size Design via Unbalanced Optimal Transport}

\author{
  Malte Franke\textsuperscript{1, 2}\thanks{Corresponding author: \texttt{malte.franke@inf.ethz.ch}} \quad
  Stefan P. Schmid\textsuperscript{1, 2} \quad
  \v{Z}arko Ivkovi\'c\textsuperscript{1, 2} \AND
  Kjell Jorner\textsuperscript{1,2} \quad
  Andreas Krause\textsuperscript{1,2} \\[1.0 em]
  \textsuperscript{1}ETH Zürich \quad
  \textsuperscript{2}NCCR Catalysis 
}

\begin{document}

\maketitle

\begin{abstract}
\looseness -1 The success of generative molecular design depends on a model's \textit{steerability} toward high-reward samples. Because many molecular properties are intrinsically linked to molecular size, accurately capturing the joint distribution of properties and the number of atoms is essential. However, current diffusion and flow-based models fix the number of atoms, which ultimately limits their ability to navigate this complex relationship. To address this, we introduce \morph, a flexible-size generative model for conditional and unconditional 3D molecular design based on geometric graphs. By dynamically adapting size, \morph can seamlessly integrate existing structural priors, like scaffolds, and significantly enhances property steering. We show that \morph matches current fixed-size state-of-the-art models while significantly increasing sampling flexibility. We demonstrate out-of-distribution generation in regimes where previous models fail, paving the way for enhanced generative modeling for molecular design.
\end{abstract}

\section{Introduction}


\looseness -1 Generative models can accelerate scientific discovery by guiding chemical space exploration, ultimately reducing the number of expensive physical experiments \cite{du2024machine}. The practical success of these models depends on i) their \textit{expressivity} — whether they can model molecular systems and their interactions,  and ii) on their \textit{steerability} — how reliably they can be guided toward high-reward regions of molecular structure distributions. These properties are essential across different modes of discovery, whether the objective is to generate entirely new molecules in de novo design, or to explore the surrounding chemical neighborhood of an initial structure to optimize for specific properties.

\looseness -1 Current diffusion and flow-based 3D generative models \cite{vignac2023midi, le2023navigating, dunn_flowmol3_2025, irwin_semlaflow_2025} face a significant algorithmic limitation: they cannot adapt the number of atoms during the generation process. Since many chemical properties depend on molecular size, this constraint artificially restricts the model's sample efficiency and limits both expressivity and steerability. To enable successful discoveries of molecules, an ideal model should learn the joint distribution of size and target properties. To this end, the model should be capable of dynamically adjusting the size (\textit{i.e.}, the number of atoms) during generation.

In this work, we introduce \morph\footnote{Code: \url{https://github.com/maltefranke/morph-chem}}\footnote{Checkpoints: \url{https://huggingface.co/maltefranke/morph}}, a flexible-size generative model for geometric graphs, building on recent progress in flexible-size sequence generation, specifically Edit Flows \cite{havasi_edit_2025}, and learning of general Markov processes via Generator Matching \cite{holderrieth_generator_2025}.
Our discrete-continuous flow-based model can dynamically change the number of nodes in a graph by performing multiple insertion and deletion operations, while substituting node and edge types and continuously moving atom positions to transform any prior geometric graph into a valid 3D molecular structure. 
Our main contributions are:
    
\begin{enumerate}
    \item The extension of flexible-size sequence generation to 3D geometric graphs, enabling the seamless integration of structural priors.

    \item A matching algorithm based on unbalanced optimal transport that aligns geometric graphs of different sizes to define the target probability paths for training.

    \item Empirical demonstration that \morph matches state-of-the-art fixed-size models in unconditional generation, while unlocking enhanced property steerability and successful out-of-distribution (OOD) generation where previous models fail.
\end{enumerate}

\section{Methods}

\subsection{Notation}
\label{sec:notation}

Let $\mathcal{A} = \{1, \dots, Z_{\text{max}}\}$ be the set of atom types and $\mathcal{B} = \{0, \dots, b_{\text{max}}\}$ be the set of bond types. We initially represent a molecule with $n$ atoms as a geometric graph $\tilde{g} = (n, \mathcal{V}, \mathcal{E})$. The indexed tuple $\mathcal{V} = (p_i)_{i=1}^n$ contains nodes $p_i = (a_i, x_i) $ with $a_i \in \mathcal{A}$ and positions $x_i \in \mathbb{R}^3$. The edges are represented by an adjacency matrix $\mathcal{E} = [e_{ij}]_{i,j=1}^n \in \mathcal{S}_n(\mathcal{B})$, where $\mathcal{S}_n(\mathcal{B})$ denotes the space of symmetric $n \times n$ bond-type matrices. 
To enforce permutation and spatial symmetries, we quotient the state space by the group $SE(3) \times S_n$, where $SE(3)$ is the Special Euclidean group acting on $x_i$, and $S_n$ is the symmetric group acting on the node indices. Because graphs with varying node counts occupy different dimensional spaces (in practice bounded by a finite maximum node count $n_{\text{max}}$), the overall space $\mathcal{G}$ is constructed as the disjoint union of these quotient spaces:
\begin{equation}
    g \in \mathcal{G} = \biguplus_{n=1}^{n_{\text{max}} }\left(\mathcal{A}^n \times \mathbb{R}^{n \times 3} \times \mathcal{S}_n(\mathcal{B}) \right) \ \big/ \ \left(SE(3) \times S_n \right)
\end{equation}
Having established the state space of geometric graphs $\mathcal{G}$\footnote{Point clouds are trivially subsumed under this formulation by setting $b_{\text{max}}=0$.}, our objective is to construct a generative process that transforms any prior graph $g_0$ into a valid molecular structure $g_1$ with a potentially different number of nodes. We can appropriately formulate these transitions as jump-flow processes which mix continuous dynamics with discrete jumps \cite{applebaum2009levy}.

\subsection{Generative Dynamics: Flow Matching and CTMC}
\label{sec:dynamics}
With fixed graph topology and dimension $n_t$, the continuous positions $x_t \in \mathbb{R}^{n_t \times 3}$ evolve under the probability flow ODE $\frac{dx_t}{dt} = v_t(g_t)$, governed by a velocity field $v_t : \mathcal{G} \to \mathbb{R}^{n_t \times 3}$, whose marginal density satisfies the continuity equation $\partial_t p_t(x) = -\nabla_x \cdot (v_t(g_t) p_t(x))$. The infinitesimal generator of these dynamics acts on test functions $f : \mathcal{G} \to \mathbb{R}$ by directional differentiation:
\begin{equation}
    [\mathcal{L}_t^{\text{flow}} f](g) \;=\; v_t(g) \cdot \nabla_x f(g).
\end{equation}
Concurrently, the graph can undergo discrete jumps via a Continuous-Time Markov Chain (CTMC) \cite{ethier2009markov}. Because these jumps, such as inserting a new atom, require proposing both discrete components (atom/bond types) and continuous 3D coordinates, we characterize transitions on $\mathcal{G}$ using a rate density $u_t(g' \mid g_t)$ defined with respect to a reference measure $\nu(dg')$. For a small time step $h$, the transition kernel from the current state $g_t$ to a measurable set $A \subseteq \mathcal{G}$ is:
\begin{equation}
    \mathbb{P}(g_{t+h} \in A \mid g_t) \;=\; \delta_{g_t}(A) \,+\, h \int_A u_t(g' \mid g_t) \, \nu(dg') \,+\, o(h),
\end{equation}
where $\delta_{g_t}(A)$ denotes the Dirac measure centered at $g_t$, ensuring the state remains unchanged unless a jump occurs into $A$. The generator of this pure-jump process defined by $u_t$ acts on test functions as:
\begin{equation}
    [\mathcal{L}_t^{\text{jump}} f](g) \;=\; \int_{\mathcal{G}} \bigl(f(g') - f(g)\bigr)\, u_t(g' \mid g)\, \nu(dg').
\end{equation}
Conceptually, the integral over the trans-dimensional space $\mathcal{G}$ with respect to $\nu(dg')$ measures the total rate of transitioning from the current graph $g$ to all possible modified graphs $g'$. Since flow and jumps act on $g_t$ independently, the generator of the coupled process is their sum:
\begin{equation}
    [\mathcal{L}_t f](g) \;=\; \underbrace{v_t(g) \cdot \nabla_x f(g)}_{\text{flow}} \;+\; \underbrace{\int_{\mathcal{G}} \bigl(f(g') - f(g)\bigr)\, u_t(g' \mid g)\, \nu(dg')}_{\text{jump}}.
    \label{eq:generator}
\end{equation}
The combined generator $\mathcal{L}_t$ governs the trans-dimensional process via the Kolmogorov forward equation $\partial_t p_t = \mathcal{L}_t^* p_t$. Thus, learning the dynamics requires fitting the drift $v_t$ and jump rates $u_t$. However, these targets rely on the intractable marginal $p_t(g)$ and are therefore inaccessible directly from data.

\subsection{Training via Conditional Generator Matching}
\label{sec:cgm}
To bypass this intractability, we utilize the Conditional Generator Matching (CGM) framework \cite{holderrieth_generator_2025}, which generalizes conditional flow matching to arbitrary Markov processes.
CGM exploits the fact that the marginal generator $\mathcal{L}_t$ defined in Eq.~\ref{eq:generator}  can be recovered as a conditional expectation of \textit{conditional} generators defined along paths between paired endpoints $(g_0, g_1) \sim \pi$, where $\pi$ is a coupling of $p_{\text{prior}}$ and $p_{\text{data}}$:
\begin{equation}
    [\mathcal{L}_t f](g) \;=\; \mathbb{E}_{(g_0, g_1) \sim \pi(\cdot, \cdot \mid g_t = g)} \!\left[\, [\mathcal{L}_t^{g_0, g_1} f](g) \,\right].
    \label{eq:cgm-generator}
\end{equation}
Because $\mathcal{L}_t$ is linear in $(v_t, u_t)$, Eq.~\ref{eq:cgm-generator} implies that a network $(v_t^\theta, u_t^\theta)$ trained to regress the conditional velocity and rates under any Bregman divergence recovers the correct marginal generator at the optimum \cite{holderrieth_generator_2025}. Learning $\mathcal{L}_t$ therefore reduces to a supervised problem over conditional paths that we construct in the following.

\subsection{Unbalanced Optimal Transport for Geometric Graphs}
\label{sec:uot}
Since $g_0$ and $g_1$ generally have different numbers of nodes (\textit{i.e.}, $n_0 \neq n_1$), before any interpolation can be defined, we must establish a structural alignment. Such an assignment should specify for each source node whether it is \textit{matched} to a target node (and thus survives, possibly with substituted type), or \textit{deleted}; and which target nodes are \textit{inserted}. 
However, finding an exact optimal transport plan between two fully connected graphs is expensive \cite{riesen2009approximate}. We bypass the combinatorial bottleneck by setting the cost of all edge edits to zero, which is justifiable for molecular geometries, where bond topology can be inferred given the 3D structure and atom valency. The resulting matching of discrete-continuous point clouds can be formulated with Unbalanced Optimal Transport (UOT) \cite{sejourne2023unbalanced}.

The UOT framework relaxes the strict mass-conservation constraints of OT, allowing us to match distributions of different masses by penalizing marginal deviations. In our case, the UOT objective seeks a transport plan $\mathbf{P} \in \mathbb{R}_{\geq 0}^{n_0 \times n_1}$ that minimizes
\begin{equation}
    \min_{\mathbf{P} \in \mathbb{R}_{\geq 0}^{n_0 \times n_1}} \left( \sum_{i=1}^{n_0} \sum_{j=1}^{n_1} \tilde{C}[i,j] P[i,j] + w_{\text{del}} ||\mathds{1}_{n_0} -\mathbf{P} \mathds{1}_{n_1}||_{1} + w_{\text{ins}} ||\mathds{1}_{n_1} -\mathbf{P}^T \mathds{1}_{n_0}||_{1} \right)
\end{equation}
where $\tilde{C}[i,j]$ are the entries of the cost matrix $\tilde{\textbf{C}} \in \mathbb{R}^{n_0\times n_1}$, and $w_{\text{del}}$ and $w_{\text{ins}}$ control the cost of mass deletion and insertion.
To obtain a set of exact graph edits, we restrict the transport plan $\mathbf{P}$ to integer assignments. This discrete UOT problem is theoretically equivalent to formulating an assignment problem in an augmented space.
By introducing auxiliary variables $\epsilon$ that lift the space to $(n_0+n_1) \times (n_0+n_1)$, the objective becomes finding the binary assignment matrix $\mathbf{\hat{P}}$ that minimizes the total cost
\begin{equation}
    \min_{\mathbf{\hat{P}} \in \{0,1\}^{(n_0+n_1) \times (n_0+n_1)}} \sum_{i=1}^{n_0+n_1} \sum_{j=1}^{n_0+n_1} \mathbf{C}[i,j] \mathbf{\hat{P}}[i,j]
    \label{eq:aug_OT}
\end{equation}
subject to the strict constraints that every row and column is assigned exactly once ($\sum_j \mathbf{\hat{P}}[i,j] = 1$ and $\sum_i \mathbf{\hat{P}}[i,j] = 1$). 
The augmented cost matrix $\mathbf{C}\in \mathbb{R}^{(n_0 + n_1)\times(n_0+n_1)}$ is structured as
\begin{equation*}
\vcenter{\hbox{\includegraphics[width=0.25\textwidth]{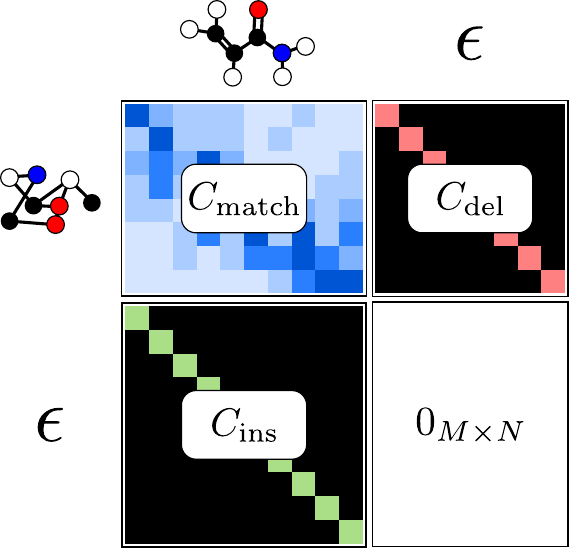}}} 
\quad \text{where:} \quad
\begin{aligned}
C_{\text{match}}[i, j] &= w_{\text{move}} \|x_{0, i} - x_{1, j}\|_2 + w_{\text{type}} \mathds{1}(a_{0, i} \neq a_{1, j}), \\
C_{\text{del}}[i, l] &= \begin{cases} w_{\text{del}} & \text{if } i+n_1=l \\ \infty & \text{otherwise} \end{cases}, \\
C_{\text{ins}}[k, j] &= \begin{cases} w_{\text{ins}} & \text{if } k=j+n_0 \\ \infty & \text{otherwise} \end{cases},
\end{aligned}
\end{equation*}

for source nodes $i \in \{1, \dots, n_0\}$, target nodes $j \in \{1, \dots, n_1\}$, and slack indices $k \in \{n_0+1, \dots, n_0+n_1\}$ and $l \in \{n_1+1, \dots, n_1+n_0\}$. 
The optimization problem in Eq. \ref{eq:aug_OT} can be efficiently solved using the Hungarian algorithm for small- to moderate sized-graphs \cite{kuhn1955hungarian, munkres1957algorithms}. However, the matching does not yet take into account the translation and rotation symmetry of the SE(3) group. Therefore, we rotationally align matched positions by applying the Kabsch algorithm \cite{kabsch1976solution} , leading to a locally optimal transport plan \cite{klein2023equivariant}. We use the resulting coupling to define interpolation paths between $g_0$ and $g_1$.

\subsection{Graph Interpolation}
\label{sec:interpolation}
We can define a conditional probability path evolving the three modalities independently as
\begin{equation}
    p_t(g \mid g_0, g_1) \;=\; p_t(x \mid x_0, x_1) \cdot p_t(a \mid a_0, a_1) \cdot p_t(e \mid e_0, e_1).
    \label{eq:cond-factorization}
\end{equation}
Under this decomposition, the conditional generator splits into a continuous velocity for surviving node coordinates and discrete rates governing four atomic operations that span all transitions in $\mathcal{G}$\footnote{Edge insertions and deletions are absorbed into edge substitution via the null bond type.}: node insertion and deletion, atom-type substitution, and edge-type substitution. 
Therefore, the optimal assignment matrix $\mathbf{\hat{P}}^*$ which aligns valid node sets $\mathcal{S}_0 = \{1, \dots, n_0\}$ from $g_0$ and $\mathcal{S}_1 = \{1, \dots, n_1\}$ from $g_1$, partitions the non-zero entries into three mutually exclusive edit sets.
Each set dictates a distinct interpolation along the time variable $t\in[0,1]$. 
Given a scheduler $\kappa_t$ with boundary conditions $\kappa_0=0$ and $\kappa_1=1$ \cite{gat_discrete_2024, campbell2024generative} we define:

\textbf{Matched nodes} $\mathcal{M} = \{ (i, j) \mid \mathbf{\hat{P}}^*[i, j] = 1, \; i \in \mathcal{S}_0, j \in \mathcal{S}_1 \}$ are present throughout the entire interpolation. Their continuous positions follow the standard Gaussian path \cite{lipman2022flow}, $x_t \sim \mathcal{N}((1-t)x_0 + tx_1, \sigma^2I)$, while atom and edge type interpolation follows the scheduler $\kappa_t^{\text{sub}}$, yielding $a_t \sim \text{Cat}\left((1-\kappa_t^{\text{sub}}) \frac{1}{|\mathcal{A}|} + \kappa_t^{\text{sub}}\delta_{a_1}\right)$ and $e_t \sim \text{Cat}\left((1-\kappa_t^{\text{sub}}) \frac{1}{|\mathcal{B}|} + \kappa_t^{\text{sub}}\delta_{e_1}\right)$.

\textbf{Insertion nodes} $\mathcal{I} = \{ (\epsilon, j) \mid \mathbf{\hat{P}}^*[k, j] = 1, \; k \notin \mathcal{S}_0, j \in \mathcal{S}_1 \}$ require sampling an independent event time $t_{\text{ins}} \sim \kappa^{\text{ins}}$ for each node. For $t \leq t_{\text{ins}}$, the node does not yet exist. If inserted ($t > t_{\text{ins}}$), its position is sampled from a Gaussian centered around the target position $x_t \sim \mathcal{N}(x_1, \sigma_t^2I)$ with a shrinking variance $\sigma_t = \sigma(1-t) + \varepsilon$. The discrete features follow the insertion scheduler $\kappa_t^{\text{ins}}$, with \\ $a_t \sim \text{Cat}\left((1-\kappa_t^{\text{ins}}) \frac{1}{|\mathcal{A}|} + \kappa_t^{\text{ins}}\delta_{a_1}\right)$ and $e_t \sim \text{Cat}\left((1-\kappa_t^{\text{ins}}) \frac{1}{|\mathcal{B}|} + \kappa_t^{\text{ins}}\delta_{e_1}\right)$.

 \textbf{Deletion nodes} $\mathcal{D} = \{ (i, \epsilon) \mid \mathbf{\hat{P}}^*[i, l] = 1, \; i \in \mathcal{S}_0, l \notin \mathcal{S}_1 \}$ are assigned an independent, node-wise deletion event time $t_{\text{del}} \sim \kappa^{\text{del}}$. For $t < t_{\text{del}}$, we fix the node and its edges at their prior attributes, i.e., position $x_t = x_0$, atom type $a_t = a_0$, and edge types $e_t = e_0$. Else, if $t \geq t_{\text{del}}$, the node and all its connected edges are removed.

\begin{figure}[h!]
    \centering
    \includegraphics[width=0.98\linewidth]{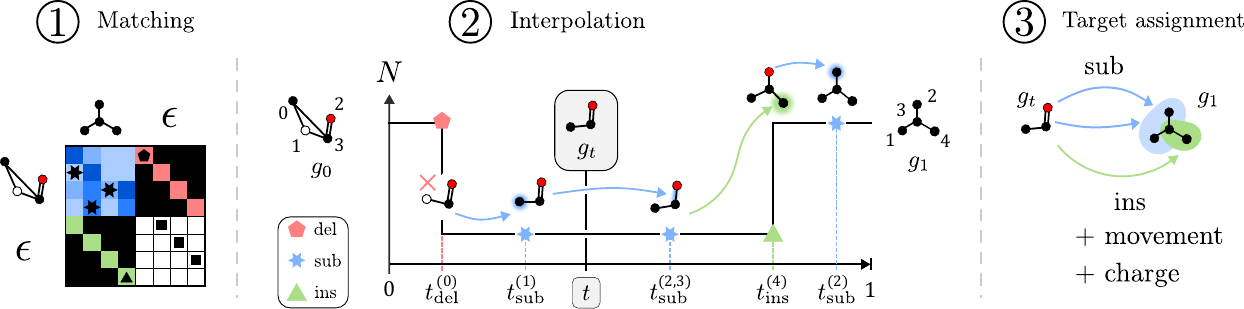}
    \caption{Interpolation between a randomly sampled graph ($t=0$) and a valid molecule ($t=1$). In the first step, we use the UOT matching to obtain edit sets determining whether nodes must be substituted, inserted or deleted. Given these sets, we interpolate to an intermediate graph $g_t$. Lastly, we assign the conditional generator matching targets.}
    \label{fig:interpolation-overview}
\end{figure}

The algorithm is shown in Algorithm \ref{alg:interp}.
Using the edit sets, the state transitions are trained via their conditional generator target using a Bregman divergence, yielding an overall objective of the form
\begin{equation}
    \mathcal{J}(\theta) \;=\; \mathbb{E}_{t,\, (g_0, g_1) \sim \pi,\, g_t \sim p_t(\cdot \mid g_0, g_1)} \!\left[\, \textstyle\sum_k \mathcal{J}_k(\theta) \,\right],
\end{equation}
 whose terms we describe in the following.

\subsection{State Transitions}
\label{sec:transitions}
All discrete structural modifications to the graph must be fully resolved by the end of the generative process at $t=1$. To achieve this, we model the transition rates of our CTMC using bounded hazard distributions. Similar to \cite{nguyen_oneflow_2025}, we choose to factorize the general transition rate $\lambda$ for any discrete action as
\begin{equation}
    \lambda = \Delta \cdot \frac{\dot{\kappa}_{t}}{1-\kappa_{t}} \geq 0.
    \label{eq:rate}
\end{equation}
In this formulation, the explicit hazard rate $\frac{\dot{\kappa}_{t}}{1-\kappa_{t}}$ determines \textit{when} the action happens. By construction, the schedule $\kappa_t \to 1$ as $t \nearrow 1$, causing the denominator $(1-\kappa_t)$ to vanish. Consequently, the transition rate $\lambda \to \infty$ forces the Markov process to resolve any remaining actions and ensures the termination of the generative process by $t=1$. 
The variable $\Delta$ determines \textit{how much} of an action happens (or if it happens at all) and is dynamically extracted from our UOT matching at time $t$.
Since we impose the scheduler $\kappa_t$, the model only has to learn the target $\Delta$. 
We now define the full set of objectives for nodes currently present in $g_t$.

\textbf{Insertions} For each insertion node ($j \in \mathcal{I}$) that has not yet been inserted ($t \leq t_{\text{ins}}^{(j)}$), we locate its nearest spatial neighbor to define $\mathcal{Q}_i = \left\{ q_j \mid j \in \mathcal{I}, \ t \le t_{\text{ins}}^{(j)}, \ i = \arg\min_{k \in g_t} \|x_k - x_j^{\mathcal{I}}\|_2 \right\}$, the set of pending insertion nodes assigned to the existing node $i$ at time $t$. The target insertion count is exactly the cardinality of this set, $\Delta^{\text{ins}}_i = |\mathcal{Q}_i|$.  We then define the insertion action as

\begin{minipage}[c]{0.25\textwidth}
    \centering
    \raisebox{-11ex}{\includegraphics[width=\linewidth]{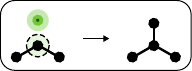}}
\end{minipage}%
\begin{minipage}[c]{0.75\textwidth}
    \begin{alignat}{2}
        &\ins(g_t, p_i, q, \mathbf{e}_q) &&= \left( n_t+1, \ \mathcal{V}_t \cup \{q\}, \, \begin{pmatrix} \mathcal{E}_t & \mathbf{e}_q \\ \mathbf{e}_q^\top & 0 \end{pmatrix}\right) \\[1ex]
        u_t^\theta(&\ins(g_t, p_i, q, \mathbf{e}_q) \mid g_t) &&= \lambda^{\text{ins}}_{t,i}(g_t) Q^{\text{ins}}_{t, i}(q \mid g_t) Q^{\text{ins}}_{t, i}(\mathbf{e}_q \mid q, g_t)
    \end{alignat}
\end{minipage}

where $q=(a, x)$ is the new node and $\mathbf{e}_q$ are its edges to all existing nodes in $g_t$ and to other inserted nodes. Importantly, the prediction of edges for inserted nodes depends on their sampled positions and types.

While in theory our CTMC formulation models single-node insertions in an infinitesimal time interval, in practice we batch the prediction of all pending nodes $\mathcal{Q} = \bigcup_{k=1}^{n_t} \mathcal{Q}_k$ during training. Therefore, we define an augmented edge target $\mathbf{e}_q^+$, which encompasses the edges connecting the newly inserted node $q$ not only to the existing graph $g_t$, but also to all other concurrently inserted nodes in $\mathcal{Q}$. 
Finally, we can write the combined loss for insertions by summing over these assigned nodes:
\begin{equation}
    \mathcal{J}_{\ins}(\theta) = \frac{1}{n_t}\sum^{n_t}_{i=1} \left[ \mathcal{J}_{\text{PoissonNLL}}(\Delta^{\theta, \text{ins}}_i, \Delta^{\text{ins}}_i) - \sum_{q \in \mathcal{Q}_i} \log Q^{\theta, \text{ins}}_{t,i}(q \mid g_t) + \sum_{q \in \mathcal{Q}_i} \mathcal{J}_{\text{CE}}(\hat{\mathbf{e}}_q^+, \mathbf{e}_q^+) \right]. 
    \label{eq:loss-ins}
\end{equation}

\textbf{Deletions} To reduce the number of nodes in a graph, we can define the deletion of node $i$ as

\begin{minipage}[c]{0.25\textwidth}
    \centering
    \includegraphics[width=\linewidth]{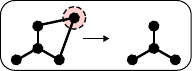}
\end{minipage}%
\begin{minipage}[c]{0.75\textwidth}
    \begin{alignat}{2}
        &\del(g_t, p_i) &&= \left(n_t-1, \ \mathcal{V}_t \setminus \{p_i\} , \ \mathcal{E}_{t, -i} \right) \\[1ex]
        u_t^\theta(&\del(g_t, p_i) \mid g_t ) &&= \lambda^{\text{del}}_{t, i}(g_t)
    \end{alignat}
\end{minipage}

where $i$ and all its edges are removed, leading to a reduced cardinality $n_t-1$.
For a node destined for deletion ($i \in \mathcal{D}$), its target is $\Delta^{\text{del}}_i = 1$ if it has not yet reached its deletion time ($t < t_{\text{del}}^{(i)}$). All other nodes have $\Delta^{\text{del}}_i=0$.
Since a node can either be deleted or not, we formulate the deletion loss as a binary cross-entropy over all present nodes
\begin{equation} 
    \mathcal{J}_{\del}(\theta) = \frac{1}{n_t}\sum^{n_t}_{i=1}\mathcal{J}_{\text{BCE}}(\Delta^{\theta, \text{del}}_i, \Delta^{\text{del}}_i).
    \label{eq:loss-del}
\end{equation} 
\textbf{Substitutions} Next, we define the substitution of node and edge types of the graph. For existing nodes, we set $\Delta^{\text{sub,a}}_i = 1$ if the atom type of a node needs to be substituted (\textit{i.e.}, $a_t^{(i)} \neq a_1^{(i)}$), and $0$ otherwise. An analogous binary target $\Delta^{\text{sub,e}}_{ij} \in \{0, 1\}$ is extracted for edges. 
We then denote the substitution of node $i$ as

\begin{minipage}[c]{0.25\textwidth}
    \centering    \includegraphics[width=\linewidth]{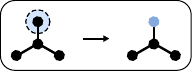}
\end{minipage}%
\begin{minipage}[c]{0.75\textwidth}
    \begin{alignat}{2}
        &\sub_{\text{a}}(g_t, p_i, p'_i) &&= \left( n_t, \ (\mathcal{V}_t \setminus \{p_i\}) \cup \{p'_i\}, \, \mathcal{E}_t \right) \\[1ex]
        u_t^\theta(&\sub_{\text{a}}(g_t, p_i, p'_i) \mid g_t) &&= \lambda^{\text{sub,a}}_{t, i}(g_t) Q^{\text{sub,a}}_{t,i}(p'_i \mid g_t).
    \end{alignat}
\end{minipage}

with $p_i' = (x_i, a_i')$. Correspondingly, the substitution of the bidirectional edge between $i$ and $j$ can be formulated as

\begin{minipage}[c]{0.25\textwidth}
    \centering \includegraphics[width=\linewidth]{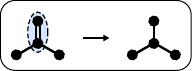}
\end{minipage}%
\begin{minipage}[c]{0.75\textwidth}
    \begin{alignat}{2}
        &\sub_{\text{e}}(g_t, e_{ij}, b) &&= \left( n_t, \ \mathcal{V}_t, \, \mathcal{E}' \right) \\[1ex]
        u_t^\theta(&\sub_{\text{e}}(g_t, e_{ij}, b) \mid g_t) &&= \lambda^{\text{sub,e}}_{t, ij}(g_t) Q^{\text{sub,e}}_{t, ij}(b \mid g_t)
    \end{alignat}
\end{minipage}

where  $\mathcal{E}'_{ij} = \mathcal{E}'_{ji} = b$ and $\mathcal{E}'_{kl} = (\mathcal{E}_t)_{kl}$ otherwise.
The substitution loss is then
\begin{align}
    \mathcal{J}_{\sub}(\theta) = \frac{1}{n_t}\sum^{n_t}_{i=1} \Bigg[ &\underbrace{\mathcal{J}_{\text{BCE}}(\Delta^{\theta,\text{sub,a}}_i, \Delta^{\text{sub,a}}_i) + \Delta^{\text{sub,a}}_i \mathcal{J}_{\text{CE}}(\hat{a}_{1,i}, a_{1,i})}_{\mathcal{J}_{\text{sub,a}}} \nonumber \\ 
    + \frac{1}{n_t}\sum_{j=1}^{n_t} &\underbrace{\mathcal{J}_{\text{BCE}}(\Delta^{\theta,\text{sub,e}}_{ij}, \Delta^{\text{sub,e}}_{ij}) + \Delta^{\text{sub,e}}_{ij} \mathcal{J}_{\text{CE}}(\hat{e}_{1, ij}, e_{1, ij})}_{\mathcal{J}_{\text{sub, e}}} \Bigg]
    \label{eq:loss-sub}
\end{align}
\textbf{Movement} Finally, the nodes can change their positions. Using endpoint targets, we define the loss only on positions of persisting nodes as

\begin{equation}
    \mathcal{J}_{\text{move}}(\theta) = \frac{1}{n_t}\sum^{n_t}_{i=1}(1- \Delta^{\text{del}}_i ) \mathcal{J}_{\text{MSE}}(\hat{x}_{1,i}, x_{1,i}). \
    \label{eq:loss-move}
\end{equation}

Although charge is not explicitly modeled as part of $g_t$, we treat its prediction as an auxiliary task. For newly inserted nodes, charge is modeled jointly with position and atom type by the insertion head, $Q_{t,i}^{\theta,\text{ins}}$. For existing nodes, we include a separate cross-entropy loss $\mathcal{J}_{\text{charge}}(\theta)$ evaluated at the final frame. Therefore, the final training objective is formulated as:
\begin{equation}
    \mathcal{J}(\theta) = \mathbb{E}_{\substack{t \sim \mathcal{U}([0, 1]),\\ (g_0, g_1) \sim \pi,\\ g_t \sim q(\cdot \mid g_0, g_1)}} \left[ \mathcal{J}_{\text{move}}(\theta) + \mathcal{J}_{\text{sub, e}}(\theta)  + \mathcal{J}_{\text{sub, a}}(\theta)  + \mathcal{J}_{\text{del}}(\theta)  + \mathcal{J}_{\text{ins}}(\theta) +\mathcal{J}_{\text{charge}}(\theta)  \right]
    \label{eq:loss-total}
\end{equation}

Given a trained model, we numerically integrate the learned jump-flow process to transform a prior graph $g_0$ into a molecule (see Alg. \ref{alg:rate-integration}). For discrete updates, we exploit the exact integrability of the schedule hazards to compute finite-step transition probabilities, with network predictions held fixed within each step. We elaborate on additional model details in \ref{apx:model_details}.

\section{Results}
\subsection{De novo design}
\label{sec:de-novo}
We show results for 3D molecular design on 
GEOM-Drugs \cite{axelrod2022geom}, targeting the unconditional design of small organic molecules. We evaluate models based on established metrics such as validity and stability (see \ref{sec:metrics}), energy, and PoseBusters \cite{buttenschoen2024posebusters}, and compare against current state-of-the-art models for fixed-size generation.
\morph achieves 95.46\% validity at 100 function evaluations (NFE), compared with 93.9\% for SemlaFlow at the same NFE.
Here, validity requires both connectivity and RDKit sanitization over all generation attempts. Joint success on all 12 PoseBusters checks is 88.30\% on GEOM-Drugs and 93.53\% on QM9, counting all attempts (Table~\ref{tab:qm9_geom_posebusters}). Additional ChEMBL3D results are reported in Appendix~\ref{sec:chembl-denovo}.

\begin{table}[H]
\centering
\small
\caption{GEOM-Drugs de novo generation using SemlaFlow metrics (3 seeds; 10k
molecules/seed). \textsuperscript{\textdagger}: adapted from SemlaFlow
\cite{irwin_semlaflow_2025}; \textsuperscript{*}: reproduced from the FlowMol3
repository. The \morph row uses the explicit-H checkpoint after 100 completed
epochs; stability uses the valence table and validity is connected sanitization.}
\begin{tabular}{cl ccccc r} 
\toprule
\multicolumn{2}{l}{Model} & Atom Stab $\uparrow$ & Mol Stab $\uparrow$ & Valid $\uparrow$ & Unique $\uparrow$ & Novel $\uparrow$  & NFE \\ \midrule
\multirow{4}{*}{\rotatebox{90}{\scriptsize $n_{\text{atoms}}$ fixed}}
& MiDi\textsuperscript{\textdagger}       & 99.8 & 91.6 & 77.8 & \textbf{100.0} & \textbf{100.0} & 500 \\
& EQGAT-diff\textsuperscript{\textdagger} & 99.8$_{\pm0.0}$ & 93.4$_{\pm0.21}$ & 94.6$_{\pm0.24}$ & \textbf{100.0}$_{\pm0.0}$ & 99.9$_{\pm0.07}$ & 500 \\
& SemlaFlow\textsuperscript{\textdagger}  & 99.8$_{\pm0.0}$ & 97.3$_{\pm0.08}$ & 93.9$_{\pm0.19}$ & \textbf{100.0}$_{\pm0.0}$ & 99.6$_{\pm0.03}$ & 100 \\ 
& FlowMol3\textsuperscript{*} & \textbf{100.0}$_{\pm0.0}$ & \textbf{99.95}$_{\pm0.02}$ & \textbf{99.35}$_{\pm0.06}$ & 99.54$_{\pm0.02}$ & – & 250 \\
\midrule

& \morph (Ours) & 99.98$_{\pm0.00}$ & 95.92$_{\pm0.33}$ & 95.46$_{\pm0.31}$ & \textbf{100.00}$_{\pm0.00}$ & 99.75$_{\pm0.07}$ & 100 \\ \bottomrule
\label{tab:geom-de-novo}
\end{tabular}
\end{table}

\begin{table}[H]
\centering
\small
\setlength{\tabcolsep}{2pt}
\caption{GEOM-Drugs molecular properties (mean$_{\pm\mathrm{SD}}$ across molecules).
Training: 248,147 molecules; \morph: 29,008 evaluated molecules pooled across
three runs; baselines: 10k/model. Bold marks the mean closest to training.}
\begin{tabular}{@{}lccccccc@{}}
\toprule
Model & \# heavy atoms & \# rot. bonds & \# HBD & \# HBA & TPSA (\AA$^2$) & \# rings & \# stereocentres \\
\midrule
GEOM \textsubscript{train} & 24.64$_{\pm5.28}$ & 4.77$_{\pm2.19}$ & 1.18$_{\pm0.89}$ & 4.47$_{\pm1.67}$ & 73.42$_{\pm27.45}$ & 2.99$_{\pm0.99}$ & 0.42$_{\pm0.92}$ \\
\midrule
SemlaFlow & 22.88$_{\pm5.03}$ & \textbf{4.67}$_{\pm2.55}$ & \textbf{1.29}$_{\pm1.01}$ & 3.93$_{\pm1.58}$ & 63.17$_{\pm26.92}$ & 2.75$_{\pm1.05}$ & 0.96$_{\pm1.36}$ \\
FlowMol3 & 22.09$_{\pm4.88}$ & 4.89$_{\pm1.93}$ & 0.93$_{\pm0.76}$ & 3.78$_{\pm1.55}$ & 60.43$_{\pm23.48}$ & 2.39$_{\pm0.94}$ & \textbf{0.36}$_{\pm0.75}$ \\
\morph (Ours) & \textbf{23.64}$_{\pm5.21}$ & 4.93$_{\pm2.33}$ & 1.40$_{\pm1.05}$ & \textbf{4.11}$_{\pm1.67}$ & \textbf{68.67}$_{\pm29.39}$ & \textbf{2.85}$_{\pm1.10}$ & 0.98$_{\pm1.34}$ \\
\bottomrule
\end{tabular}
\label{tab:geom-de-novo-properties}
\end{table}

Furthermore, \morph most closely matches the training means for four of seven
medicinally relevant descriptors: heavy-atom count, hydrogen-bond acceptors,
rings, and topological polar surface area (Table~\ref{tab:geom-de-novo-properties}).
SemlaFlow is closest for rotatable bonds and hydrogen-bond donors, and FlowMol3
for stereocentres.

Together with the stability and validity results above, the distributional metrics highlight competitive sample quality in addition to \morph's more expressive generation trajectories.
Next, we demonstrate the steerability of our model on conditional design tasks.

\begin{figure}[ht!]
  \centering
  \begin{subfigure}[b]{0.32\textwidth}
    \centering
    \includegraphics[width=\textwidth]{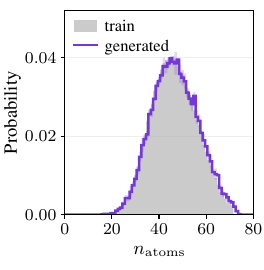}
    \caption{Empirical size targets}
    \label{fig:OOD_in_distr}
  \end{subfigure}
  \hfill
  \begin{subfigure}[b]{0.32\textwidth}
    \centering
    \includegraphics[width=\textwidth]{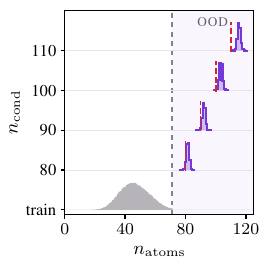}
    \caption{OOD size targets}
    \label{fig:OOD_cond}
  \end{subfigure}
  \hfill
  \begin{subfigure}[b]{0.32\textwidth}
    \centering
    \includegraphics[width=\textwidth]{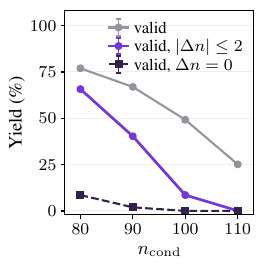}
    \caption{Valid molecule yield}
    \label{fig:OOD_validity}
  \end{subfigure}
  \caption{Size conditioning on GEOM-Drugs (training cutoff: 72 atoms).
  Yields use all generation attempts; validity requires a connected, sanitizable
  molecule, and $\Delta n=n_{\mathrm{atoms}}-n_{\mathrm{cond}}$.
  We generate 10k attempts per seed for empirical targets and 5k per fixed target;
  bands and error bars show standard deviations across three seeds.}
  \label{fig:OOD_results}
\end{figure}

\paragraph{Property steering}
To show the steerability of \morph in an illustrative setting, we generate
structures of a desired size given an uninformative prior on the number of
atoms $n_0\sim\mathcal{U}(n_{\text{min}},n_{\text{max}})$. We train a model
conditioned on the target atom count on GEOM-Drugs \cite{axelrod2022geom}, with
training molecules capped at 72 atoms including hydrogens. After validating
in-distribution steerability with empirical atom-count targets
(Fig.~\ref{fig:OOD_in_distr}), we request $n_{\mathrm{cond}}\in\{80,90,100,110\}$,
beyond the training cutoff.

\looseness -1 Figure~\ref{fig:OOD_results} shows that \morph recovers the training size
distribution with 93.84\% validity for empirical targets and can be steered to generate larger molecules under OOD conditioning. At $n_{\mathrm{cond}}=80$, validity is 76.96\%
and the atom-count MAE is 1.59. As targets increase, generated sizes increasingly
overshoot them: at $n_{\mathrm{cond}}=110$, validity falls to 25.14\% and MAE rises
to 5.38 atoms. The joint yield of valid molecules within two atoms of the target,
and especially at the exact target, therefore decreases
(Fig.~\ref{fig:OOD_validity}). Full results are in Table~\ref{tab:geom-n-updated}.

\paragraph{Towards out-of-distribution discovery}
\looseness -1 To demonstrate \morph's potential for OOD discovery, we generate molecules with a desired log partition coefficient (logP), which is a measure of lipophilicity with strong correlation to size. We use RDKit \cite{landrum2026rdkit} to calculate logP values of all QM9 molecules and finetune a property-conditioned model.

\begin{figure}[H]
    \centering
    \includegraphics[width=0.98\linewidth]{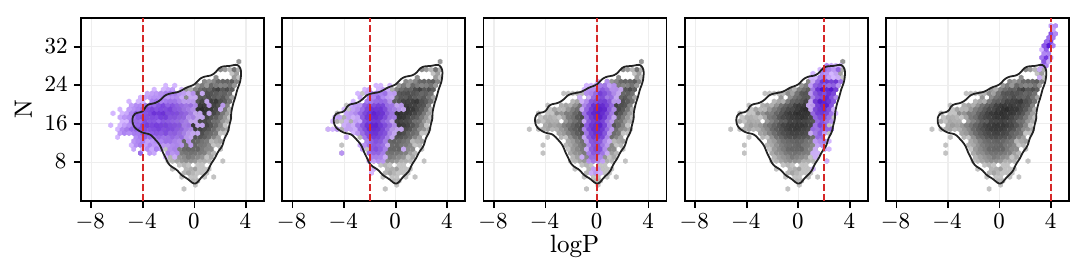}
    \caption{LogP-conditioned generation with \morph (10k samples). Gray contours
    show the training density, purple hexagons generated molecules, and dashed red lines
    the target logP. Despite uniform sampling of $n_0$,
    $p(n_1\mid\text{logP})$ follows the corresponding training slice.}
    \label{fig:logp_results}
\end{figure}

Figure \ref{fig:logp_results} shows that our model can steer size depending on the property conditioning, automatically recovering the corresponding training data $n$-marginal distribution.
When conditioned on values in low support or OOD regions ($\text{logP}=-4$ and $4$, respectively), we observe that \morph can successfully generate valid molecules with sizes and logP values outside the training distribution. We validate multi-property steerability in \ref{sec:apx_logp_qed}. Pooled logP evaluations on QM9 and GEOM-Drugs achieve MAEs of 0.304 and 0.538, respectively (Table~\ref{tab:logp-updated}) without externally supplying the target-dependent size distribution.
Since scientific discovery requires generating candidates with properties unseen in the training data, \morph's flexible-size molecule generation can lead to improved property-conditioned generative molecular design.

\subsection{Scaffold decoration}
\label{sec:scaffold_dec}
\looseness -1 Besides improved property steering, \morph's flexible-size generation also enables the
construction of molecules based on a given structural prior. Such a task is known as
\textit{scaffold decoration}, or lead optimization, and is common in drug discovery to
optimize a structure while preserving the key molecular properties tied to its scaffold
\cite{fialkova_2022, Schneuing2024-by}.

\noindent
\begin{minipage}{\linewidth}
\begin{minipage}[t]{0.64\linewidth}
\vspace{0pt}
\setlength{\parskip}{0.4em}
\looseness -1 We demonstrate this task on GEOM-Drugs, starting from Bemis--Murcko scaffolds $g_0$ and
extending them to complete molecules $g_1$ that contain the same motif
(Fig.~\ref{fig:scaffold_gen}). To construct the training paths, we first identify the
common scaffold between $g_0$ and $g_1$. Its atom and bond types remain fixed throughout
the trajectory. We then apply UOT independently at each substituent site, building
substituents through atom insertions, substitutions, and movement.

The test evaluation uses 512 scaffolds with eight completions each (4,096 generation attempts per seed, three seeds). \morph achieves
$99.99 \pm 0.00\%$ atom stability, $97.90 \pm 0.16\%$ molecule stability,
$97.99 \pm 0.20\%$ validity, $63.54 \pm 0.59\%$ pooled uniqueness, and
$97.52 \pm 0.10\%$ novelty. All eight native PoseBusters checks pass for $94.10 \pm 0.19\%$ of all attempts (Table~\ref{tab:geom_scaffold_decoration}).
\end{minipage}\hfill
\begin{minipage}[t]{0.32\linewidth}
\vspace{0pt}
\centering
\includegraphics[width=0.96\linewidth]{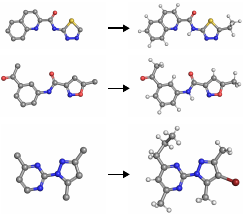}
\captionof{figure}{Scaffold priors (left) and \morph completions (right).}
\label{fig:scaffold_gen}
\end{minipage}
\end{minipage}
\par

We additionally fine-tune the ChEMBL3D implicit-H \morph model on the in-pocket CrossDocked2020 multi-R benchmark \cite{francoeur20203d} from DiffDec \cite{Xie_2024_DiffDec}. Given only an initial scaffold and protein pocket, \morph generates multiple substituents simultaneously without anchor-atom annotations or a prescribed final number of atoms. It produces 89.80\% valid molecules, and 40.33\% of its completions have higher predicted binding affinity than their reference molecules, yielding competitive performance against specialized in-pocket methods (Table~\ref{tab:scaffold-pocket-baselines}).

\section{Related Work}

\paragraph{Flexible-size generative models}

Generative modeling of flexible-size data has been predominantly explored for discrete sequences.  Autoregressive (AR) models naturally increase sequence length with strict ordering (\textit{e.g.}, left-to-right), but generally prohibit retrospective editing. To alleviate this, several variable-length sequence models based on insertions and deletions have been developed \cite{gu2019levenshteintransformer, ruis2020insertiondeletiontransformer, johnson2021inplacecorruptioninsertiondeletion}. Recent advances in masked diffusion models enable any-order sequence modeling with token insertion and re-masking capabilities \cite{kim2025anyorderflexiblelengthmasked, patel2025insertion}. Flexible flow-based generative models like Edit Flows \cite{havasi_edit_2025} can model substitution, insertion, and deletion of tokens, even supporting multi-modal insertions \cite{nguyen_oneflow_2025}.

For graphs, AR methods sequentially add nodes and edges \cite{you2018graphrnn, liao2019efficient}, but inherit the same ordering limitations as sequence models. Diffusion models \cite{vignac2022digress, jo2022score} generate graphs by evolving node and edge types simultaneously on a fixed-dimensional state space. To enable variable-size generation, they rely on a predetermined maximum size and utilize padding nodes or masking rather than actively scaling the dimensionality during generation. Only recently has this limitation been addressed by reformulating discrete graph diffusion to explicitly support either insertion or deletion of nodes  \cite{ninniri2025graphdiffusioninsertdelete}.

While frameworks for flexible-size generation exist in discrete domains, translating these transdimensional operations to discrete-continuous \textit{geometric} graphs—where nodes possess continuous 3D coordinates and are subject to SE(3) symmetries—remains an open problem.

\paragraph{Flexible-size generative models for 3D molecular design}
Most diffusion and flow-based generative models for 3D molecular design decouple the determination of system size from the generation process itself. A common approach is to draw $n$ from an empirical distribution, \textit{i.e.}, $n_0 \sim p_{\text{data}}(n)$, and keep it fixed along the generation \cite{hoogeboom2022equivariant, vignac2023midi, le2023navigating, irwin_semlaflow_2025, joshi_all-atom_2025, zeng_propmolflow_2026, morehead_zatom-1_2026, Schneuing2024-by, Xie_2024_DiffDec, vonessen_tabasco_2025, reidenbach2026applications}. Other methods predict $n_0$ via a conditional task beforehand \cite{igashov2024equivariant, zhou_paflow_2025}, or make use of 'fake atoms' \cite{dunn_flowmol3_2025}. Still, none of the listed methods go beyond the empirical distribution of number of atoms seen during training.

\looseness -1 To overcome fixed-size limitations, several works have explored dynamic dimensionality during generation, though they face \textit{directional} and hence geometric constraints. Transdimensional jump-diffusion \cite{campbell_trans-dimensional_2023} and AR models \cite{cheng_scalable_2025, Joshi_2021_3DScaffold, daigavane2024symphony, rose2025neat} are restricted to \textit{insertions}, \textit{i.e.} they cannot transform a graph $g_0$ into another graph $g_1$ if its number of nodes $n_0 $ is larger than $n_1$. 
Lifting \textit{directional} constraints remains largely untackled, limiting the use of informative structural priors. 

\looseness -1 Other approaches increase flexibility by allowing both insertions and deletions. Branching Flows \cite{billera_branching_2025} constructs a tree-based generative process via in-place duplication and coalescence of atoms using the CGM framework, but relies on the molecule's sequence representations to define training targets. 
On the other hand, by using different representations, fragment-based flow models \cite{poletukhin_3d_2026} can alter the number of atoms during the trajectory by changing fragment types. Beyond graph-based representations, other paradigms bypass discrete nodes entirely by modeling molecules in continuous spaces, such as 3D voxels \cite{pinheiro2024structurebaseddrugdesigndenoising, pinheiro_3d_2024, faltings_proxelgen_2025} or continuous fields \cite{veljkovic_cords_2026}.

\section{Limitations}
\label{sec:limitations}
The jump process introduces separate objectives for insertions, deletions, and
substitutions, making optimization and loss balancing more complex than in a
fixed-dimensional flow. We defer systematic ablation of the interpolation
schedules, matching procedure and -costs, or relative loss weights to future work. Although \morph samples
quickly at low NFE, validity depends on the integration budget
(Appendix~\ref{sec:sampling-efficiency}), resulting in a trade-off between speed and
sample quality.

Furthermore, our extrapolation experiments remain simple.
Atom-count steering on GEOM-Drugs shows declining validity and size accuracy
beyond training support. 
Reliability under arbitrary distribution shifts
remains to be tested. Broader evaluations should cover chemically meaningful
properties and more diverse molecular settings.

\section{Conclusion}
In this work, we introduced \morph, a flexible-size 3D generative model for geometric graphs which can simultaneously perform \textit{multiple} insertions, deletions and substitutions. We show competitive performance to leading fixed-size generative models on de novo design tasks while matching the training distribution better and being significantly more expressive. \morph also enables flexible generation based on structural priors, as validated on a scaffold decoration task. By being able to generate OOD samples, \morph is a promising building block for accelerating discoveries in the chemical sciences and beyond.

\begin{ack}
This publication was created as part of NCCR Catalysis (grant numbers 180544 and 225147), a National Centre of Competence in Research funded by the Swiss National Science Foundation.
This work was supported as part of the Swiss AI Initiative by a grant from the Swiss National Supercomputing Centre (CSCS) under project ID a131 on Alps.
\end{ack}

\bibliographystyle{unsrtnat}
\bibliography{literature}


\appendix
\raggedbottom

\section{Technical Appendices and Supplementary Material}

\subsection{Data details}

\paragraph{QM9} The QM9 dataset \cite{ramakrishnan2014quantum} contains around 134k small organic molecules with up to 9 heavy atoms of types  \{ \texttt{H, C, F, N, O} \}. It is a subset of the enumerated GDB-17 dataset \cite{ruddigkeit2012enumeration}, filtered with heuristic for chemical plausibility. Structures were generated at the B3LYP/6-31G(2df,p) level of theory.
There have been several inconsistencies in earlier versions of QM9, including molecules with non-zero charges and invalid bond orders.
Therefore, we choose to use the corrected QM9 dataset from \cite{zeng_propmolflow_2026}, and follow their train, validation and test split.


\paragraph{GEOM-Drugs} The GEOM-Drugs dataset consists of more than 240k molecules relevant for computer-aided drug design \cite{axelrod2022geom}. Each molecule is represented by multiple conformers, leading to over 5.5M 3D structures.
It contains atom types \{ \texttt{B, Bi, Br, C, Cl, F, H, I, N, O, P, S, Si} \}.
Similar to SemlaFlow \cite{irwin_semlaflow_2025}, to increase training efficiency, we decide to remove all molecules with more than 72 atoms which constitute about 1\% of the training data, and follow their train-val-test split. 

\subsection{Choices for schedules}
\label{sec:schedules}
When designing our schedules, we considered different stages across the generation time. In the first stage the model should focus on generating the rough molecule geometry and topology, whereas for late $t$ the structure is finalized and relaxed to its energy minimum. To realize this, we make use of Smoothstep functions of the type
\begin{equation}
    \kappa_t(\alpha) = 3t^{2 \alpha} - 2t^{3\alpha}
\end{equation}
which fulfill $\kappa_{t=0}(\alpha)=0$ and $\kappa_{t=1}(\alpha)=1$.
We choose early-peaking schedules with $\alpha=0.8$ for insertions and deletions. Substitutions are chosen to (likely) resolve after insertions ($\alpha=1.5$), such that erroneous insertions could be corrected in the remaining time-interval.

\subsection{Model details}
\label{apx:model_details}
\begin{figure}[!htb]
    \centering
    \includegraphics[width=\linewidth]{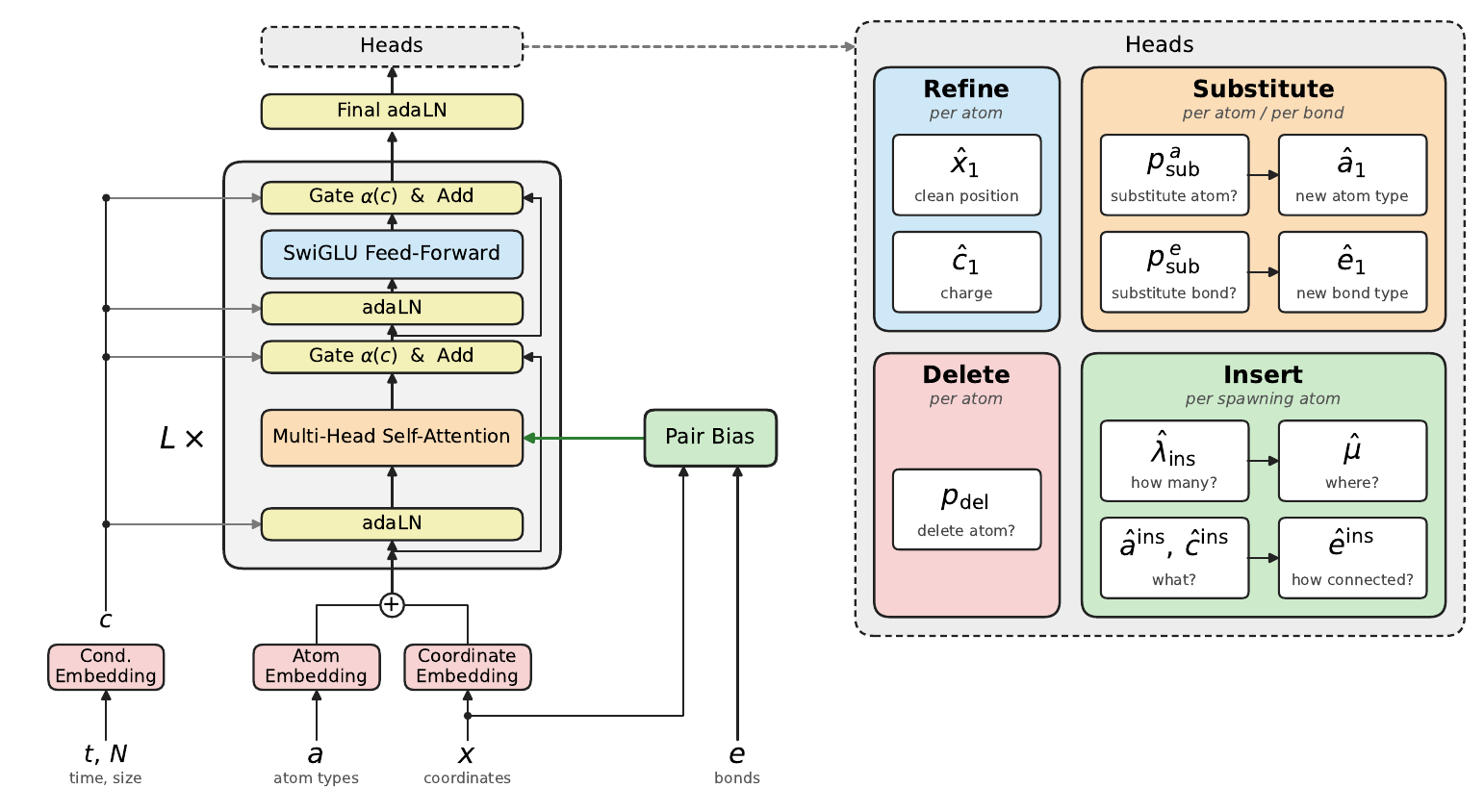}
    \caption{Architecture of \morph. Left: the DiT backbone combines atom and
    coordinate embeddings with time- and atom-count conditioning and pairwise
    attention biases. Right: readout heads predict clean coordinates and charges,
    substitutions, deletions, and unimodal insertions, including their count,
    position, atom type, charge, and bonds.}
    \label{fig:model-architecture}
\end{figure}

We use a Diffusion Transformer (DiT) backbone similar to \cite{joshi_all-atom_2025} with about 41M parameters for all experiments. The DiT-based model learns approximate equivariance via rotation augmentations. We therefore supply one random rotation for each molecule per batch.
Figure~\ref{fig:model-architecture} summarizes the backbone and output heads.

\paragraph{Embeddings and backbone}
Atom and bond types use learned categorical embeddings. We encode time with
sinusoidal features and the current atom count with low-frequency Fourier
features concatenated with a normalized scalar count. The latter retains a
continuous count input beyond the training range. A learned projection of
centered, normalized coordinates is added to the atom embeddings.
The backbone uses eight transformer blocks of width 512 with 16 attention
heads, query/key normalization, and SwiGLU feed-forward layers. Time and count
embeddings condition the blocks through adaLN-Zero. Attention biases encode
pairwise distances, bond types, and atom-type pairs.
The output node and pair features have dimension 256.

\paragraph{State and edit heads}
Small neural readouts predict atom types, auxiliary charges, bond types, and
edit decisions. The insertion-count head uses softplus; deletion and
substitution logits are mapped to probabilities with a sigmoid.
A residual coordinate head predicts clean positions and removes their
graph-wise center of mass. Existing-bond features combine the sum and absolute
difference of projected endpoint features with bond and distance embeddings,
giving symmetric bond predictions.

\paragraph{Unimodal insertion head}
Each spawning node $i$, with hidden features $h_i$ and position $x_i$, predicts
one clean insertion position and categorical atom-type and charge probabilities:
\begin{equation}
    \mu_i=x_i+f_x(h_i),\qquad
    A_i=\operatorname{softmax}(f_a(h_i)),\qquad
    C_i=\operatorname{softmax}(f_c(h_i)).
\end{equation}
The readouts $f_x,f_a,f_c$ use separate normalized, gated adapters and linear
output projections. At inference, the clean insertion position is $\mu_i$,
while atom type and charge are sampled from $A_i$ and $C_i$. The prescribed
interpolation noise supplies the positional spread.

\paragraph{Inserted bonds}
For a new atom $q$ spawned by $i$ and an existing atom $j$, an MLP predicts
bond logits from
\begin{equation}
    m_{qj}=\big[h_i\parallel h_j\parallel a_j\parallel a_q\parallel c_q
    \parallel\operatorname{RBF}(d_{qj})\parallel\operatorname{RBF}(d_{qi})\big],
\end{equation}
where atom types and charges are one-hot encoded, $d_{qj}$ and $d_{qi}$ are
distances from the clean inserted position, and $\parallel$ denotes
concatenation. For bonds between two inserted atoms, a shared encoder combines
each atom's spawning-node features, predicted attributes, and distance to its
spawn. A second MLP uses the sum and absolute difference of these endpoint
encodings together with their interatomic distance, ensuring symmetric logits.
Bonds are predicted from clean inserted attributes; positions, atom types,
and bonds are then noised to the next integration time using the interpolation
schedules. Charge remains an auxiliary clean prediction.

\paragraph{Practical deviations from the full objective}
The implementation uses a simplified insertion-content objective.
The Poisson count loss still supervises the full pending count
$\Delta_i^{\mathrm{ins}}=|\mathcal{Q}_i|$, but the unimodal head is trained only
on the nearest pending atom assigned to each spawning node with a nonempty
$\mathcal{Q}_i$. Its position uses mean squared error over the three spatial
coordinates, and its atom type and charge use cross-entropy.
This replaces the likelihood sum over all assigned atoms in the full
insertion-loss expression. Bond supervision still covers all pending
inserted atoms, including bonds to existing atoms and between inserted atoms,
using their clean target attributes. Components are averaged over their
supervised targets within each molecule.

For practical inference, we cap insertions at one atom per spawning node per
integration step. We combine the predicted pending count with the integrated
schedule probability, clip the result to $[0,1]$, and draw a Bernoulli event.
This cap applies to sampling; the training count target includes all pending
insertions. Further atoms can be inserted in subsequent steps as the model
is evaluated on the updated graph.

\subsection{Training details}
\label{sec:train_details}
We optimize hidden two-dimensional weight matrices with Muon and use auxiliary
AdamW for the remaining parameters, including embeddings and prediction heads.
Tables~\ref{tab:hyperparameters} and~\ref{tab:loss_weights} report the settings
shared by the released unconditional QM9, GEOM-Drugs, and ChEMBL3D models.
Both optimizers use linear warmup followed by cosine decay. We train in BF16
mixed precision with $t\sim\mathcal{U}(0,1)$ and evaluate the exponential
moving average (EMA) of the model parameters.

\begin{table}[H]
\begin{minipage}[t]{0.47\linewidth}
\vspace{0pt}
\centering
\caption{Optimizer and scheduler settings. The schedule horizon includes warmup.}
\label{tab:hyperparameters}
\small
\setlength{\tabcolsep}{3pt}
\begin{tabular}{@{}lc@{}}
\toprule
\textbf{Parameter} & \textbf{Value} \\
\midrule
Muon peak learning rate & $10^{-2}$ \\
Muon momentum (Nesterov) & 0.95 \\
Newton--Schulz iterations & 5 \\
Muon $\varepsilon$ & $10^{-7}$ \\
\midrule
AdamW peak learning rate & $3\times10^{-4}$ \\
AdamW $(\beta_1,\beta_2)$ & $(0.9,0.95)$ \\
AdamW $\varepsilon$ & $10^{-10}$ \\
\midrule
Weight decay (both) & 0 \\
Global gradient-norm clip & 1.0 \\
LR warmup steps & 2,000 \\
LR schedule horizon (steps) & 200,000 \\
Minimum / peak learning rate & 0.05 \\
EMA decay & 0.999 \\
EMA warmup steps & 5,000 \\
\bottomrule
\end{tabular}
\end{minipage}\hfill
\begin{minipage}[t]{0.50\linewidth}
\vspace{0pt}
\centering
\caption{Components of Eq.~\eqref{eq:loss-total}, with references to their definitions.}
\label{tab:loss_weights}
\small
\setlength{\tabcolsep}{3pt}
\begin{tabular}{@{}llc@{}}
\toprule
\textbf{Objective} & \textbf{Component} & \textbf{Weight} \\
\midrule
$\mathcal{J}_{\text{move}}$~\eqref{eq:loss-move} & MSE & 2.00 \\ 
$\mathcal{J}_{\text{charge}}$~\eqref{eq:loss-total} & CE & 0.50 \\ 
\multirow{2}{*}{$\mathcal{J}_{\text{sub,a}}$~\eqref{eq:loss-sub}}
 & Decision (BCE) & 0.20 \\ 
 & Type (CE) & 0.03 \\ 
\multirow{2}{*}{$\mathcal{J}_{\text{sub,e}}$~\eqref{eq:loss-sub}}
 & Decision (BCE) & 0.25 \\ 
 & Type (CE) & 0.50 \\ 
$\mathcal{J}_{\text{del}}$~\eqref{eq:loss-del} & BCE & 0.40 \\ 
\multirow{4}{*}{$\mathcal{J}_{\text{ins}}$~\eqref{eq:loss-ins}}
 & Count (Poisson NLL) & 0.15 \\ 
 & Content (approx.) & 0.01 \\ 
 & Bonds to existing & 0.40 \\ 
 & Bonds to inserted & 0.03 \\ 
\bottomrule
\end{tabular}
\end{minipage}
\end{table}

The implemented objective sums component losses averaged over supervised
targets per molecule and then over molecules with targets. Component weights
are fixed, categorical class weights are uniform, and no additional time
weighting is applied. The insertion-content weight multiplies the sum of
position MSE and atom-type and charge cross-entropies. Appendix~\ref{apx:model_details}
describes the practical deviations from the full objective.

The released unconditional models were trained on two NVIDIA RTX 4090 GPUs. QM9 uses 1,024 molecules per GPU and two gradient-accumulation
steps, giving an effective batch size of 4,096. GEOM-Drugs and both ChEMBL3D
variants use 320 molecules per GPU without accumulation, giving an effective
batch size of 640. The reported checkpoints correspond to 1,500 completed
epochs for QM9, 100 for GEOM-Drugs, and 200 for both ChEMBL3D variants.
Checkpoint configurations are distributed with the public code and model release.

The fixed-target property-conditioned QM9 models were trained on four NVIDIA
RTX 4090 GPUs with an effective batch size of 1,024 for 500 epochs
(<12h total runtime).

\subsection{Metrics}
\label{sec:metrics}
We use the following metric definitions, with stability evaluated according to
SemlaFlow \cite{irwin_semlaflow_2025}:
\begin{itemize}
    \item \textbf{Atom stability} is the fraction of atoms satisfying the element- and formal-charge-dependent valence table, including implicit hydrogens.
    \item \textbf{Molecule stability} is the fraction of generation attempts in which every atom satisfies this table; failed molecule construction counts as unstable.
    \item \textbf{Validity} is the fraction of all generation attempts that produce a connected graph passing RDKit sanitization (\texttt{connected-sanitization-validity}), with no additional radical or net-charge filter.
    \item \textbf{Uniqueness} is the number of distinct canonical molecular identities divided by the number of SMILES-convertible outputs. Updated summaries use hydrogen-stripped identities.
    \item \textbf{Novelty} is the fraction of SMILES-convertible outputs whose canonical identity is absent from the training reference, with duplicates retained in the denominator.
\end{itemize}

For all reported PoseBusters evaluations, pass
rates divide the number of passing structures by the total number of generation
attempts. Failed molecule construction and unavailable checks count as failures, and
joint success requires every enabled check to pass. The unconditional, logP, and atom-count
evaluations use all 12 checks, including internal energy and flatness, whereas the native
scaffold protocol uses eight checks and excludes energy and flatness. Unless stated otherwise, uncertainties are sample standard deviations across three seeds, including uncertainties on mean properties. The multi-property experiment instead reports within-run property spreads.

\subsection{Interpolation algorithm}
\begingroup
\setlength{\intextsep}{4pt}

\begin{algorithm}[H]
\footnotesize
\caption{Discrete Interpolation}
\label{alg:interp-discrete}
    \begin{algorithmic}
        \Function{InterpolateDiscrete}{$y_0, y_1, \kappa$}
        \State Sample $u \sim \mathcal{U}(0, 1)$
        \State \Return $y_0$ if $u > \kappa$ else $y_1$
    \EndFunction
    \end{algorithmic}
\end{algorithm}

\begin{algorithm}[H]
\caption{Molecular Graph Interpolation (Flow Matching Training)}
\label{alg:interp}
\small
\begin{algorithmic}[1]
    \Require Prior sample $g_0 \sim \pi$ with nodes $\mathcal{S}_0$, Target graph $g_1 \sim p_{\mathrm{data}}$ with nodes $\mathcal{S}_1$, Time $t \in [0, 1)$
    \Ensure Interpolated state $g_t$, Active targets $g_1^{\mathrm{filt}}$, Pending insertions $\mathcal{Q}$
    
    \Statex \hspace{1em}\textit{// 1. Optimal Transport Alignment}
    \State Compute optimal assignment matrix $\mathbf{\hat{P}}^*$ aligning $\mathcal{S}_0$ and $\mathcal{S}_1$ \Comment{Align valid node sets}
    \State Define node partitions based on non-zero entries of $\mathbf{\hat{P}}^*$:
    \Statex \quad $\mathcal{M} \gets \{i \mid \mathbf{\hat{P}}^*[i, j] = 1, i \in \mathcal{S}_0, j \in \mathcal{S}_1\}$ \Comment{Matched nodes}
    \Statex \quad $\mathcal{D} \gets \{i \mid \mathbf{\hat{P}}^*[i, l] = 1, i \in \mathcal{S}_0, l \notin \mathcal{S}_1\}$ \Comment{Deletion nodes}
    \Statex \quad $\mathcal{I} \gets \{j \mid \mathbf{\hat{P}}^*[k, j] = 1, k \notin \mathcal{S}_0, j \in \mathcal{S}_1\}$ \Comment{Insertion nodes}
    
    \Statex \hspace{1em}\textit{// 2. Evaluate Schedules and Sample Event Times}
    \State Compute schedulers $\kappa^{\mathrm{sub}}_t, \kappa^{\mathrm{del}}_t, \kappa^{\mathrm{ins}}_t$
    \State Sample independent deletion event times $t_{\mathrm{del}, i} \sim \kappa^{\mathrm{del}}$ for $i \in \mathcal{D}$
    \State Sample independent insertion event times $t_{\mathrm{ins}, j} \sim \kappa^{\mathrm{ins}}$ for $j \in \mathcal{I}$
    
    \Statex \hspace{1em}\textit{// 3. Determine Active Nodes at Time $t$}
    \State $\mathcal{V}_t \gets \mathcal{M}$
    \State $\mathcal{V}_t \gets \mathcal{V}_t \cup \{i \in \mathcal{D} \mid t < t_{\mathrm{del}, i}\}$ \Comment{Not yet deleted}
    \State $\mathcal{V}_t \gets \mathcal{V}_t \cup \{j \in \mathcal{I} \mid t > t_{\mathrm{ins}, j}\}$ \Comment{Already inserted}
    \State Let $n_t \gets |\mathcal{V}_t|$
    
    \Statex \hspace{1em}\textit{// 4. Interpolate Graph Features}
    \For{each active node $k \in \mathcal{V}_t$}
        \If{$k \in \mathcal{D}$}
            \State $x_{t,k} \gets x_{0,k}$ \Comment{Fix positions at prior}
            \State $a_{t,k} \gets a_{0,k}$ \Comment{Fix atom types at prior}
        \ElsIf{$k \in \mathcal{I}$}
            \State $\sigma_t \gets \sigma(1-t) + \varepsilon$
            \State Sample $x_{t,k} \sim \mathcal{N}(x_{1,k}, \sigma_t^2 I)$ \Comment{Gaussian centered on target}
            \State $a_{t,k} \gets \mathrm{InterpolateDiscrete}(a_{0,k}, a_{1,k}, \kappa^{\mathrm{ins}}_t)$
        \Else \Comment{$k \in \mathcal{M}$}
            \State Sample $x_{t,k} \sim \mathcal{N}((1-t)x_{0,k} + t x_{1,k}, \sigma^2 I)$ \Comment{Standard Gaussian path}
            \State $a_{t,k} \gets \mathrm{InterpolateDiscrete}(a_{0,k}, a_{1,k}, \kappa^{\mathrm{sub}}_t)$
        \EndIf
    \EndFor
    
    \For{each valid pair $(u, v)$ in $\mathcal{V}_t$}
        \If{$u \in \mathcal{M} \land v \in \mathcal{M}$}
            \State $e_{t,uv} \gets \mathrm{InterpolateDiscrete}(e_{0,uv}, e_{1,uv}, \kappa^{\mathrm{sub}}_t)$
        \ElsIf{$u \in \mathcal{I} \lor v \in \mathcal{I}$}
            \State $e_{t,uv} \gets \mathrm{InterpolateDiscrete}(e_{0,uv}, e_{1,uv}, \kappa^{\mathrm{ins}}_t)$
        \Else \Comment{Involves deletion nodes}
            \State $e_{t,uv} \gets e_{0,uv}$ \Comment{Fix edges at prior attributes}
        \EndIf
    \EndFor
    \State Symmetrize $\mathcal{E}_t$ so $e_{t,vu} = e_{t,uv}$
    
    \Statex \hspace{1em}\textit{// 5. Construct Targets and Local Spawn Assignments}
    \State $g_t \gets (n_t, \mathcal{V}_t, \mathcal{E}_t)$
    \State Filter $\tilde{g}_1$ to only contain target features for active nodes $\mathcal{V}_t$, yielding $g_1^{\mathrm{filt}}$
    \State Initialize pending insertion sets $\mathcal{Q}_k \gets \emptyset$ for all $k \in \mathcal{V}_t$
    \For{each pending insertion node $j \in \mathcal{I} \setminus \mathcal{V}_t$}
        \State $k^{\ast} \gets \arg\min_{k \in \mathcal{V}_t} \| x_{t,k} - x_{1,j} \|_2$ \Comment{Find nearest active neighbor}
        \State $\mathcal{Q}_{k^{\ast}} \gets \mathcal{Q}_{k^{\ast}} \cup \{j\}$
    \EndFor
    
    \State Center $x_t$ and active targets $x_1^{\mathrm{filt}}$ to zero center-of-mass
    \State \Return $g_t, g_1^{\mathrm{filt}}, \{\mathcal{Q}_k\}_{k=1}^{n_t}$
\end{algorithmic}
\end{algorithm}
\endgroup
\clearpage

\subsection{Sampling}
\begin{algorithm}[H]
\caption{Jump-flow integration for molecular generation}
\label{alg:rate-integration}
\small
\begin{algorithmic}[1]
\Require Hazard schedules $\kappa^{\mathrm{del}}_t, \kappa^{\mathrm{sub}}_t, \kappa^{\mathrm{ins}}_t$; time grid $0 = t_0 < t_1 < \dots < t_S = 1$ with $\Delta t_s = t_{s+1}-t_s$; prior $\pi$; trained network $u_t^\theta$
\State Sample node count $n_0 \sim \mathcal{U}(\mathcal{N}_{\mathrm{train}})$ \Comment{Uniform draw over train distribution of atom counts}
\State Sample prior graph $g_0 = (n_0, \mathcal{V}_0, \mathcal{E}_0) \sim \pi(g_0 \mid n_0)$ \Comment{$\mathcal{V}_0 = (p_i)_{i=1}^{n_0}$ and $\mathcal{E}_0 = [e_{ij}]_{i,j=1}^{n_0}$}
\For{$s = 0, \dots, S-1$}
    \State $t \gets t_s$, \quad $\Delta t \gets \Delta t_s$
    \State $H^k_{t,\Delta t} \gets \displaystyle\int_t^{t+\Delta t} \frac{\dot{\kappa}^k_\tau}{1-\kappa^k_\tau}\,d\tau = \log\frac{1-\kappa^k_t}{1-\kappa^k_{t+\Delta t}}$ for $k \in \{\mathrm{del}, \mathrm{sub}, \mathrm{ins}\}$
    \State $\rho^k_{t,\Delta t} \gets 1-\exp(-H^k_{t,\Delta t}) = \frac{\kappa^k_{t+\Delta t}-\kappa^k_t}{1-\kappa^k_t}$ \Comment{Conditional schedule probability; equals $1$ at $t+\Delta t=1$}
    \State $\big( (\hat{x}_{i,1}, \hat{a}_{i,1}, \hat{c}_{i,1})_{i=1}^{n_t}, [\hat{e}_{ij,1}]_{i,j=1}^{n_t}, \Delta^{\mathrm{sub,a}}, \Delta^{\mathrm{sub,e}}, \Delta^{\mathrm{del}}, \Delta^{\mathrm{ins}}, Q^{\mathrm{ins}}_{t} \big) \gets u_t^\theta(g_t)$ \\
    \Statex \hspace{1em}\textit{// 1. Continuous updates}
    \For{each node index $i \in \{1, \dots, n_t\}$}
        \State $x_i \gets x_i + \frac{\hat{x}_{i,1} - x_i}{1 - t}\,\Delta t$ \Comment{Position Euler step}
        \State $c_i \gets \hat{c}_{i,1}$ \Comment{Auxiliary charge prediction takes final-frame target}
    \EndFor \\
    \Statex \hspace{1em}\textit{// 2. Node Deletions and Substitutions (Competing Events)}
    \For{each node index $i \in \{1, \dots, n_t\}$}
        \State $q^{\mathrm{sub,a}}_i \gets \Delta^{\mathrm{sub,a}}_i \cdot \rho^{\mathrm{sub}}_{t,\Delta t}$
        \State $q^{\mathrm{del}}_i \gets \Delta^{\mathrm{del}}_i \cdot \rho^{\mathrm{del}}_{t,\Delta t}$
        \State $q^{\mathrm{total}}_i \gets q^{\mathrm{sub,a}}_i + q^{\mathrm{del}}_i$
        \State Sample $u_1, u_2 \sim \mathcal{U}(0,1)$
        \If{$u_1 < q^{\mathrm{total}}_i$}
            \If{$u_2 < q^{\mathrm{del}}_i / q^{\mathrm{total}}_i$}
                \State \textbf{Remove} $p_i$ and its incident edges from $g_t$ \Comment{Node deletion}
            \Else
                \State $a_i \gets \hat{a}_{i,1}$ \Comment{Atom-type substitution}
            \EndIf
        \EndIf
    \EndFor \\
    \Statex \hspace{1em}\textit{// 3. Edge Substitutions}
    \For{each valid pair $(i, j)$ with $i < j$ in surviving graph}
        \State Sample $b^{\mathrm{sub,e}}_{ij} \sim \mathrm{Bernoulli}(\Delta^{\mathrm{sub,e}}_{ij} \cdot \rho^{\mathrm{sub}}_{t,\Delta t})$
        \If{$b^{\mathrm{sub,e}}_{ij} = 1$}
            \State $e_{ij} \gets \hat{e}_{ij,1}$
        \EndIf
    \EndFor
    \State Symmetrize $\mathcal{E}_t$ so $e_{ji} = e_{ij}$ \\
    \Statex \hspace{1em}\textit{// 4. Node Insertions}
    \For{each node index $i$ present at start of step}
        \State Sample $b^{\mathrm{ins}}_i \sim \mathrm{Bernoulli}(\min\{1, \Delta^{\mathrm{ins}}_i \cdot \rho^{\mathrm{ins}}_{t,\Delta t}\})$ \Comment{At most one insertion per spawning node}
        \If{$b^{\mathrm{ins}}_i = 1$}
            \State Sample $(x_q, a_q, c_q) \sim Q^{\mathrm{ins}}_{t,i}(q \mid g_t)$ \Comment{Unimodal insertion head}
            \State Sample edges $\mathbf{e}_q \sim Q^{\mathrm{ins}}_{t,i}(\mathbf{e}_q \mid q, g_t)$
            \State Add new node $p_q = (a_q, x_q)$, charge $c_q$, and edges $\mathbf{e}_q$ to $g_t$
        \EndIf
    \EndFor
    \State Update $n_{t+\Delta t} \gets |\mathcal{V}_{t+\Delta t}|$
\EndFor
\State \Return $g_1$
\end{algorithmic}
\end{algorithm}
\clearpage

\captionsetup{font=small}
\setlength{\intextsep}{6pt plus 2pt minus 2pt}
\section{Further results}

\subsection{De novo design}
\begin{table}[h]
\centering
\small
\caption{QM9 de novo generation. Baselines are adapted from SemlaFlow
\cite{irwin_semlaflow_2025}. The \morph row uses the explicit-H checkpoint after
1,500 completed epochs, evaluated with 10,000 attempts per seed and three seeds.
Stability uses the valence table; validity is connected sanitization.}
\begin{tabular}{cl cccc r} 
\toprule
\multicolumn{2}{l}{Model} & Atom Stab $\uparrow$ & Mol Stab $\uparrow$ & Valid $\uparrow$ & Unique $\uparrow$ & NFE \\ \midrule

\multirow{4}{*}{\rotatebox{90}{\scriptsize $n_{\text{atoms}}$ fixed}}
& FlowMol    & 99.7 & 96.2 & 97.3 & --    & 100 \\
& MiDi       & 99.8 & 97.5 & 97.9 & 97.6  & 500 \\
& EQGAT-diff & 99.9$_{\pm0.0}$ & 98.7$_{\pm0.18}$ & 99.0$_{\pm0.16}$ & \textbf{100.0}$_{\pm0.0}$ & 500\\
& SemlaFlow  & 99.9$_{\pm0.0}$ & \textbf{99.7}$_{\pm0.03}$ & \textbf{99.4}$_{\pm0.03}$ & 95.4$_{\pm0.0}$  & 100 \\ \midrule

& \morph (Ours) & \textbf{100.00}$_{\pm0.00}$ & 99.11$_{\pm0.06}$ & \underline{99.14}$_{\pm0.08}$ & 94.91$_{\pm0.11}$ & 100 \\ \bottomrule
\label{tab:qm9-de-novo}
\end{tabular}
\end{table}

\begin{table}[H]
\centering
\footnotesize
\setlength{\tabcolsep}{3pt}
\caption{Unconditional PoseBusters results. Every value is a percentage of all
10,000 generation attempts per seed, reported as mean$_{\pm\mathrm{SD}}$ across
3 seeds. Failed construction and unavailable checks count as failures.
QM9 and GEOM-Drugs use explicit-H checkpoints after 1,500 and 100 completed epochs.
Both ChEMBL3D models use checkpoints after 200 epochs. All 12 checks are
enabled; the implicit-H ChEMBL3D scores refer to heavy-atom structures.}
\begin{tabular}{lcccc}
\toprule
Metric & QM9 & GEOM-Drugs & \shortstack{ChEMBL3D\\explicit H} & \shortstack{ChEMBL3D\\implicit H} \\
\midrule
Molecule loaded & 99.16$_{\pm0.08}$ & 96.69$_{\pm0.22}$ & 96.83$_{\pm0.13}$ & 99.05$_{\pm0.18}$ \\
Sanitization & 99.16$_{\pm0.08}$ & 96.69$_{\pm0.22}$ & 96.83$_{\pm0.13}$ & 99.05$_{\pm0.18}$ \\
All atoms connected & 99.14$_{\pm0.08}$ & 95.46$_{\pm0.31}$ & 95.80$_{\pm0.17}$ & 98.87$_{\pm0.18}$ \\
Bond lengths & 99.15$_{\pm0.08}$ & 96.04$_{\pm0.25}$ & 96.33$_{\pm0.18}$ & 97.33$_{\pm0.06}$ \\
Bond angles & 99.15$_{\pm0.09}$ & 96.41$_{\pm0.22}$ & 96.57$_{\pm0.15}$ & 98.58$_{\pm0.17}$ \\
Internal steric clash & 99.01$_{\pm0.11}$ & 92.53$_{\pm0.10}$ & 95.95$_{\pm0.12}$ & 98.31$_{\pm0.24}$ \\
No radicals & 99.01$_{\pm0.05}$ & 94.10$_{\pm0.31}$ & 94.44$_{\pm0.11}$ & 99.05$_{\pm0.18}$ \\
InChI convertible & 99.15$_{\pm0.08}$ & 96.67$_{\pm0.24}$ & 96.81$_{\pm0.13}$ & 99.05$_{\pm0.18}$ \\
Internal energy & 94.21$_{\pm0.17}$ & 96.38$_{\pm0.25}$ & 96.55$_{\pm0.17}$ & 98.78$_{\pm0.15}$ \\
Aromatic ring flatness & 99.16$_{\pm0.08}$ & 96.69$_{\pm0.23}$ & 96.82$_{\pm0.12}$ & 99.04$_{\pm0.18}$ \\
Double-bond flatness & 98.82$_{\pm0.12}$ & 96.25$_{\pm0.34}$ & 95.88$_{\pm0.14}$ & 98.59$_{\pm0.22}$ \\
Non-aromatic ring non-flatness & 98.89$_{\pm0.10}$ & 96.16$_{\pm0.15}$ & 96.40$_{\pm0.11}$ & 98.55$_{\pm0.21}$ \\
\midrule
All 12 checks & 93.53$_{\pm0.19}$ & 88.30$_{\pm0.24}$ & 91.46$_{\pm0.24}$ & 95.27$_{\pm0.16}$ \\
\bottomrule
\end{tabular}
\label{tab:qm9_geom_posebusters}
\end{table}

\subsection{Additional de novo experiment on ChEMBL3D}
\label{sec:chembl-denovo}
We additionally evaluate unconditional generation on ChEMBL3D using the processed
LoQI release \cite{nikitin2025scalable}, retaining its train/validation/test
partitions. The released training subset is restricted to molecules with at most
100 atoms before removing hydrogens. We evaluate both explicit-H graphs and
heavy-atom graphs with implicit hydrogens.

For each hydrogen representation, we select the checkpoint with the highest
mean connected-sanitization validity among the three supplied checkpoints. Both
selected models have completed 200 training epochs. This selection criterion
follows the paper's definition of validity.
Each seed generates 10,000 structures. Joint PoseBusters success
uses all attempts and requires all 12 enabled checks; individual pass rates are
reported in Table~\ref{tab:qm9_geom_posebusters}.

\begin{table}[H]
\begin{minipage}[c]{0.44\linewidth}
\captionsetup{justification=raggedright,singlelinecheck=false}
\caption{Unconditional ChEMBL3D generation with explicit-H and implicit-H models after 200
completed epochs. Rates are percentages, except diversity, and are reported as
mean$_{\pm\mathrm{SD}}$ across three seeds. Validity and PoseBusters use all
generation attempts. Uniqueness and novelty use SMILES-convertible outputs.}
\label{tab:chembl-denovo}
\end{minipage}\hfill
\begin{minipage}[c]{0.54\linewidth}
\centering
\footnotesize
\setlength{\tabcolsep}{2.5pt}
\begin{tabular}{lcc}
\toprule
Metric & Explicit H & Implicit H \\
\midrule
Atom stability & 99.98$_{\pm0.00}$ & 99.99$_{\pm0.00}$ \\
Molecule stability & 96.14$_{\pm0.18}$ & 98.85$_{\pm0.21}$ \\
Connected-sanitization validity & 95.80$_{\pm0.17}$ & 98.87$_{\pm0.18}$ \\
All 12 PoseBusters checks & 91.46$_{\pm0.24}$ & 95.27$_{\pm0.16}$ \\
Uniqueness & 99.99$_{\pm0.02}$ & 100.00$_{\pm0.01}$ \\
Novelty & 99.69$_{\pm0.05}$ & 99.64$_{\pm0.10}$ \\
Diversity & 0.8953$_{\pm0.0002}$ & 0.9140$_{\pm0.0002}$ \\
\bottomrule
\end{tabular}
\end{minipage}
\end{table}

\subsection{Sampling efficiency}
\label{sec:sampling-efficiency}
This separate GEOM-Drugs timing experiment was run on one NVIDIA RTX 4090
primarily to compare sampling speed under shared hardware. Validity values
therefore need not exactly match the aggregated results in
Table~\ref{tab:geom-de-novo}. At the same 100 NFE,
\morph generates 27.6 molecules/s with 95.6\% validity, compared with 9.2
molecules/s and 93.9\% validity for SemlaFlow. Thus, \morph provides three times
the throughput at this operating point. Varying its NFE exposes a controllable
speed--quality trade-off, from 202.5 molecules/s at 10 NFE to 97.0\% validity at
200 NFE. FlowMol3 attains the highest validity, but uses 250 NFE and samples 2.7
molecules/s.

\begin{table}[H]
\centering
\small
\caption{Sampling efficiency on one NVIDIA RTX 4090. The bold \morph row
(100 NFE) is the setting used throughout the main text.}
\begin{tabular}{lrrr}
\toprule
Model & NFE & Throughput (molecules/s) & Validity (\%) \\
\midrule
SemlaFlow & 100 & 9.2 & 93.9 \\
FlowMol3 & 250 & 2.7 & 99.3 \\
\midrule
\multirow{7}{*}{\morph (Ours)}
& 10 & 202.5 & 88.1 \\
& 25 & 83.0 & 94.0 \\
& 50 & 46.1 & 95.7 \\
& \textbf{100} & \textbf{27.6} & \textbf{95.6} \\
& 200 & 15.3 & 97.0 \\
& 500 & 6.6 & 96.3 \\
& 1000 & 3.4 & 96.6 \\
\bottomrule
\end{tabular}
\label{tab:sampling-efficiency}
\end{table}

\begin{figure}[H]
\centering
\begin{tikzpicture}[x=4.5cm, y=0.34cm, font=\footnotesize]
    \foreach \x/\tick in {0.077/3, 0.600/10, 1.077/30, 1.600/100, 1.901/200} {
        \draw[gray!20] (\x,0) -- (\x,12);
        \draw (\x,0) -- ++(0,-0.16) node[below] {\tick};
    }
    \foreach \y/\tick in {0/88, 2/90, 4/92, 6/94, 8/96, 10/98, 12/100} {
        \draw[gray!20] (0,\y) -- (2,\y);
        \draw (0,\y) -- ++(-0.025,0) node[left] {\tick};
    }
    \draw[->] (0,0) -- (2.04,0);
    \draw[->] (0,0) -- (0,12.35);
    \node[below=6mm] at (1,0) {Throughput (molecules/s; log scale)};
    \node[rotate=90, above=8mm] at (0,6) {Validity (\%)};

    \draw[GalacticPurple, thick]
        plot coordinates {(1.906,0.1) (1.519,6.0) (1.264,7.7) (1.041,7.6)
                          (0.785,9.0) (0.420,8.3) (0.132,8.6)};
    \foreach \x/\y/\nfe/\where in {
        1.906/0.1/10/above right,
        1.519/6.0/25/above right,
        1.264/7.7/50/above right,
        1.041/7.6/100/below,
        0.785/9.0/200/above,
        0.420/8.3/500/below,
        0.132/8.6/1000/above right} {
        \filldraw[fill=white, draw=GalacticPurple, line width=0.8pt]
            (\x,\y) circle[radius=1.7pt] node[\where, text=GalacticPurple,
            font=\scriptsize] {\nfe};
    }

    \node[draw=black, fill=black, circle, minimum size=3.4pt, inner sep=0pt]
        at (0.564,5.9) {};
    \node[anchor=north west, font=\scriptsize] at (0.575,5.8) {SemlaFlow};
    \node[draw=SupernovaGold, fill=SupernovaGold, minimum size=4pt, inner sep=0pt]
        at (0.031,11.3) {};
    \node[anchor=west, font=\scriptsize] at (0.045,11.3) {FlowMol3};
    \node[anchor=east, text=GalacticPurple, font=\scriptsize] at (1.98,11.4)
        {\morph labels denote NFE};
\end{tikzpicture}
\caption{Validity--throughput trade-off on one NVIDIA RTX 4090. The line connects
\morph operating points; numerical labels give NFE.}
\label{fig:sampling-efficiency}
\end{figure}

\subsection{Scaffold decoration on GEOM-Drugs}

\begin{table}[H]
\small
\begin{minipage}[c]{0.50\linewidth}
\captionsetup{justification=raggedright,singlelinecheck=false}
\caption{GEOM-Drugs scaffold decoration with the rank-16 LoRA model
after 25 completed fine-tuning epochs. Each seed generates eight completions for
each of 512 test scaffolds (4,096 attempts). Rates are percentages of all attempts,
reported as mean$_{\pm\mathrm{SD}}$ across three seeds; failed construction and
unavailable checks count as failures. The native eight-check protocol excludes
energy and flatness and uses no inherited-clash exemptions.}
\label{tab:geom_scaffold_decoration}
\end{minipage}\hfill
\begin{minipage}[c]{0.46\linewidth}
\centering
\begin{tabular}{@{}lc@{}}
\toprule
Metric & Pass rate (\%, $\uparrow$) \\
\midrule
Molecule loaded & 98.13$_{\pm0.15}$ \\
Sanitization & 98.13$_{\pm0.15}$ \\
All atoms connected & 97.99$_{\pm0.20}$ \\
Bond lengths & 97.29$_{\pm0.22}$ \\
Bond angles & 97.79$_{\pm0.12}$ \\
Internal steric clash & 95.64$_{\pm0.22}$ \\
No radicals & 97.44$_{\pm0.17}$ \\
InChI convertible & 98.13$_{\pm0.15}$ \\
\midrule
All eight checks & 94.10$_{\pm0.19}$ \\
\bottomrule
\end{tabular}
\end{minipage}
\end{table}

\begin{samepage}
\subsection{In-pocket scaffold decoration}

For in-pocket scaffold decoration, we initialize \morph from the pretrained
ChEMBL3D implicit-H model and fine-tune it on the CrossDocked2020 multi-R task
used by DiffDec \cite{Xie_2024_DiffDec}. Given an
initial scaffold in a protein pocket, the model generates multiple synthetically
relevant substituents simultaneously, without information about anchor atoms or
the final number of atoms. We take the baseline results directly from DiffDec.
Because DiffDec uses QVina whereas we use AutoDock Vina 1.2.7, we report the
reference score from each evaluation so that generated molecules can be compared
with their corresponding reference molecules.\par
\end{samepage}

\begin{table}[H]
\centering
\footnotesize
\setlength{\tabcolsep}{2.25pt}
\caption{In-pocket scaffold decoration on the CrossDocked2020 multi-R task.
Baseline results, including QVina scores, are reported from DiffDec
\cite{Xie_2024_DiffDec}; \morph and its reference molecules are evaluated with
AutoDock Vina 1.2.7. Docking scores should therefore be compared with the
corresponding reference row rather than directly across the two blocks. High
affinity is the percentage of generated molecules with a lower docking score
than their corresponding reference molecule. The best comparable generated-model
result in each column is shown in bold.}
\begin{tabular}{@{}lrrrrrr@{}}
\toprule
Method
& \shortstack{Valid.\\(\%, $\uparrow$)}
& \shortstack{Unique\\(\%, $\uparrow$)}
& \shortstack{Recovery\\(\%, $\uparrow$)}
& \shortstack{Similarity\\($\uparrow$)}
& \shortstack{Docking score\\(kcal/mol, $\downarrow$)}
& \shortstack{High affinity\\(\%, $\uparrow$)} \\
\midrule
Scaffolds & -- & -- & -- & -- & $-7.77$ & -- \\
DiffDec reference molecules & -- & -- & -- & -- & $-8.46$ & -- \\
\midrule
LibINVENT & 8.62 & 6.93 & 26.85 & 0.70 & $-7.83$ & 4.57 \\
Pocket2Mol & 60.90 & 50.84 & 1.34 & 0.68 & $-8.04$ & 31.32 \\
FLAG & 88.18 & 64.73 & 0.00 & 0.62 & $-7.73$ & 38.62 \\
DiffDec (w/o anchors) & 89.84 & \textbf{71.75} & 29.54 & 0.73 & $-8.15$ & 34.67 \\
DiffDec & \textbf{97.13} & 60.12 & \textbf{45.34} & \textbf{0.80} & $-8.19$ & \textbf{40.50} \\
\midrule
\morph reference molecules & -- & -- & -- & -- & $-8.42$ & -- \\
\morph (Ours) & 89.80 & 66.43 & 37.83 & 0.77 & $-8.09$ & 40.33 \\
\bottomrule
\end{tabular}
\label{tab:scaffold-pocket-baselines}
\end{table}

\begin{figure}[H]
    \centering
    \includegraphics[width=\linewidth]{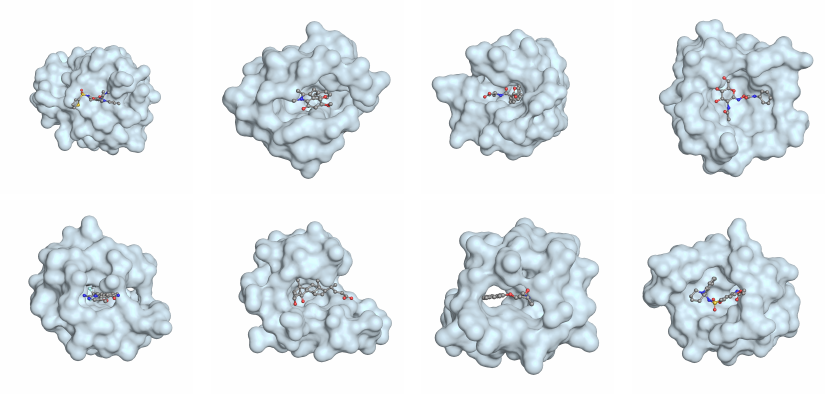}
    \caption{Example in-pocket scaffold decorations generated by \morph on the
    CrossDocked2020 multi-R task. Protein pockets are rendered as surfaces and
    generated molecules as sticks.}
    \label{fig:protein-scaffold-decoration}
\end{figure}

Using the standard \morph formulation, without anchor annotations or the final
number of atoms, we obtain competitive results on in-pocket scaffold decoration.
In particular, 89.80\% of generated molecules are valid and 40.33\% have higher
predicted binding affinity than their corresponding reference molecule.
Qualitative examples are shown in Fig.~\ref{fig:protein-scaffold-decoration}.

\begin{samepage}
\subsection{Atom-count conditioning on GEOM-Drugs}
\label{sec:geom-n-updated}
The empirical-target evaluation uses 10,000 attempts per seed, and each fixed target
uses 5,000 attempts per seed. All runs use three seeds. The generated
atom-count mean, MAE, and exact-match rate include all attempts, rather than only
valid outputs. The fixed targets exceed the 72-atom training cutoff, therefore operating in the OOD regime.
Validity and exact size targeting deteriorate as the requested size increases.\par
\end{samepage}

\begin{table}[H]
\centering
\footnotesize
\setlength{\tabcolsep}{3pt}
\caption{GEOM-Drugs atom-count conditioning. Validity uses connected
sanitization; PB is joint success on all 12 PoseBusters checks. Both rates and
exact matches are percentages of all attempts. Values are mean$_{\pm\mathrm{SD}}$
across three seeds. Uniqueness is 100\% for every setting; novelty is
99.90$_{\pm0.03}$\% for empirical targets and 100\% for fixed targets.}
\begin{tabular}{@{}l@{\;}cccccc@{}}
\toprule
\multicolumn{2}{l}{Target $n$} & Valid (\%) & PB (\%) & Mean $n$ & MAE (atoms) & Exact (\%) \\
\midrule
\multicolumn{2}{l}{Empirical ($n<72$)} & 93.84$_{\pm0.09}$ & 85.55$_{\pm0.45}$ & 46.87$_{\pm0.06}$ & 0.790$_{\pm0.003}$ & 37.21$_{\pm0.10}$ \\
\addlinespace[2pt]
\ldelim\{{4}{*}[OOD] & 80 & 76.96$_{\pm0.09}$ & 61.60$_{\pm0.51}$ & 81.56$_{\pm0.01}$ & 1.590$_{\pm0.006}$ & 10.65$_{\pm0.26}$ \\
& 90 & 66.93$_{\pm0.60}$ & 45.53$_{\pm0.93}$ & 92.29$_{\pm0.01}$ & 2.294$_{\pm0.014}$ & 2.50$_{\pm0.19}$ \\
& 100 & 49.20$_{\pm1.15}$ & 25.72$_{\pm0.44}$ & 103.47$_{\pm0.01}$ & 3.465$_{\pm0.009}$ & 0.05$_{\pm0.03}$ \\
& 110 & 25.14$_{\pm0.73}$ & 7.71$_{\pm0.20}$ & 115.38$_{\pm0.02}$ & 5.378$_{\pm0.021}$ & 0.00$_{\pm0.00}$ \\
\bottomrule
\end{tabular}
\label{tab:geom-n-updated}
\end{table}

\subsection{Pooled logP conditioning}
\label{sec:logp-updated}
The pooled evaluations use 10,000 generation attempts per dataset and seed
on QM9 and GEOM-Drugs. They are separate from the fixed-target QM9
experiments below. Validity and PoseBusters success use all attempts. LogP MAE
and $R^2$ use outputs with a finite computed logP; the scored fraction makes
this population explicit. The target sets have mean (standard deviation)
0.213 (0.889) on QM9 and 2.811 (1.329) on GEOM-Drugs.

\begin{table}[H]
\centering
\footnotesize
\setlength{\tabcolsep}{3pt}
\caption{Pooled logP-conditioned generation. Validity is connected
sanitization, PB denotes success on all 12 PoseBusters checks, and scored is the
fraction with a finite computed logP. These three fractions are percentages of
all generation attempts. Values are mean$_{\pm\mathrm{SD}}$ across three seeds.}
\begin{tabular}{lccccc}
\toprule
Dataset & Valid (\%) & PB (\%) & Scored (\%) & LogP MAE & $R^2$ \\
\midrule
QM9 & 98.69$_{\pm0.05}$ & 92.83$_{\pm0.18}$ & 98.73$_{\pm0.05}$ & 0.304$_{\pm0.002}$ & 0.815$_{\pm0.003}$ \\
GEOM-Drugs & 94.49$_{\pm0.05}$ & 87.14$_{\pm0.21}$ & 95.96$_{\pm0.14}$ & 0.538$_{\pm0.001}$ & 0.728$_{\pm0.002}$ \\
\bottomrule
\end{tabular}
\label{tab:logp-updated}
\end{table}

\begin{table}[H]
\centering
\footnotesize
\setlength{\tabcolsep}{1.8pt}
\caption{PoseBusters for conditional generation. Every value is a percentage
of all 10,000 attempts per run, including failed construction and unavailable
checks. Pooled evaluations report mean$_{\pm\mathrm{SD}}$ across three seeds;
GEOM-$n$ uses empirical atom-count targets. Fixed-target QM9 results use one run
per target, matching Table~\ref{tab:cfg_merged_logp}. All columns use the same
12-check protocol, and joint success requires all 12 checks to pass.}
\begin{tabular}{@{}lccc@{\hspace{7pt}}ccccc@{}}
\toprule
& \multicolumn{3}{c}{Pooled / empirical targets}
& \multicolumn{5}{c}{Fixed-target QM9 logP} \\
\cmidrule(lr){2-4}\cmidrule(lr){5-9}
Metric & QM9 logP & GEOM logP & GEOM-$n$ & $-4$ & $-2$ & $0$ & $2$ & $4$ \\
\midrule
Molecule loaded & 98.73$_{\pm0.05}$ & 95.96$_{\pm0.14}$ & 95.51$_{\pm0.13}$ & 77.30 & 96.33 & 98.99 & 97.85 & 31.57 \\
Sanitization & 98.73$_{\pm0.05}$ & 95.96$_{\pm0.14}$ & 95.51$_{\pm0.13}$ & 77.30 & 96.33 & 98.99 & 97.85 & 31.57 \\
All atoms connected & 98.69$_{\pm0.05}$ & 94.49$_{\pm0.05}$ & 93.84$_{\pm0.09}$ & 76.82 & 96.20 & 98.95 & 97.72 & 27.55 \\
Bond lengths & 98.69$_{\pm0.04}$ & 95.16$_{\pm0.13}$ & 94.73$_{\pm0.23}$ & 76.95 & 96.25 & 98.89 & 97.78 & 30.76 \\
Bond angles & 98.72$_{\pm0.06}$ & 95.70$_{\pm0.20}$ & 95.18$_{\pm0.15}$ & 77.14 & 96.27 & 98.95 & 97.83 & 31.38 \\
Internal steric clash & 98.62$_{\pm0.04}$ & 91.90$_{\pm0.14}$ & 91.00$_{\pm0.12}$ & 76.13 & 95.65 & 98.86 & 97.79 & 30.61 \\
No radicals & 98.23$_{\pm0.06}$ & 93.05$_{\pm0.13}$ & 91.86$_{\pm0.35}$ & 65.13 & 94.42 & 98.42 & 97.43 & 27.30 \\
InChI convertible & 98.73$_{\pm0.05}$ & 95.94$_{\pm0.14}$ & 95.46$_{\pm0.12}$ & 76.10 & 96.28 & 98.99 & 97.85 & 31.57 \\
Internal energy & 93.66$_{\pm0.24}$ & 95.62$_{\pm0.19}$ & 95.16$_{\pm0.08}$ & 75.03 & 91.67 & 93.57 & 92.90 & 31.40 \\
Aromatic ring flatness & 98.73$_{\pm0.05}$ & 95.96$_{\pm0.14}$ & 95.50$_{\pm0.14}$ & 77.30 & 96.31 & 98.99 & 97.85 & 31.57 \\
Double-bond flatness & 98.48$_{\pm0.05}$ & 95.57$_{\pm0.14}$ & 94.98$_{\pm0.08}$ & 77.20 & 96.24 & 98.59 & 97.20 & 31.57 \\
Non-aromatic ring non-flatness & 98.57$_{\pm0.03}$ & 95.56$_{\pm0.19}$ & 94.94$_{\pm0.13}$ & 76.88 & 96.16 & 98.73 & 97.67 & 31.57 \\
\midrule
All 12 checks & 92.83$_{\pm0.18}$ & 87.14$_{\pm0.21}$ & 85.55$_{\pm0.45}$ & 62.17 & 88.94 & 92.50 & 92.02 & 26.93 \\
\bottomrule
\end{tabular}
\label{tab:conditional-posebusters}
\label{tab:qm9-logp-posebusters}
\end{table}

\subsection{Fixed-target logP conditioning on QM9}
\label{sec:apx_logp}
\begin{table}[H]
\footnotesize
\caption{Fixed-target logP conditioning on QM9 (10,000 generation attempts per
target, one run per target). Rates are percentages. Validity requires connected
sanitization; PB is joint success on all 12 PoseBusters checks over all attempts.
Stability uses the valence table. $N_{\text{logP}}$ counts outputs with finite logP;
the mean$_{\pm\mathrm{SD}}$ and MAE use all such outputs.
These runs are distinct from the pooled evaluation in
Table~\ref{tab:logp-updated}; Table~\ref{tab:logp-baselines} reports the same
\morph runs under stricter molecule filtering.}
\centering
\setlength{\tabcolsep}{3pt}
\begin{tabular}{@{}cccccccccc@{}}
\toprule
logP & Valid & PB & Uniq. & Nov. & \shortstack{Atom\\stab.} & \shortstack{Mol.\\stab.} & $N_{\text{logP}}$ & $\mu_{\text{logP}}$ & $\text{MAE}_{\text{logP}}$ \\
\midrule
-4 & 76.82 & 62.17 & 65.81 & 81.55 & 99.63 & 72.91 & 7,730 & -2.833$_{\pm0.976}$ & 1.271 \\
-2 & 96.20 & 88.94 & 68.63 & 51.02 & 99.97 & 95.82 & 9,633 & -1.654$_{\pm0.468}$ & 0.465 \\
0 & 98.95 & 92.50 & 90.87 & 23.50 & 99.99 & 98.88 & 9,899 & 0.077$_{\pm0.371}$ & 0.305 \\
2 & 97.72 & 92.02 & 62.97 & 11.06 & 99.99 & 97.66 & 9,785 & 2.039$_{\pm0.383}$ & 0.313 \\
4 & 27.55 & 26.93 & 17.39 & 98.80 & 99.55 & 27.96 & 3,157 & 3.751$_{\pm0.151}$ & 0.261 \\
\bottomrule
\end{tabular}
\label{tab:cfg_merged_logp}
\end{table}

\begin{samepage}
Pooling the 36,845 sanitized, connected, neutral, radical-free \morph outputs
with finite logP across targets gives a global MAE/$R^2$ of 0.468/0.916.
SemlaFlow reports 0.607/0.909 using $p(n)$ and 0.489/0.943 using the oracle
$p(n \mid \text{logP})$ (Table~\ref{tab:logp-baselines}). The different molecule
filters limit direct comparison of these scores.\par
\end{samepage}

\begin{table}[H]
\begin{minipage}[c]{0.38\linewidth}
\captionsetup{font=small,justification=raggedright,singlelinecheck=false}
\caption{Target-wise logP-steering performance using 10,000 generation
attempts per target for \morph (one seed) and 5,000 per target for
SemlaFlow. \morph statistics use sanitized, connected, neutral,
radical-free molecules; the existing SemlaFlow results use sanitized
molecules without the same additional filters.
SemlaFlow selects its fixed atom count from either $p(n)$ or the oracle
$p(n \mid \text{logP})$ estimated on the training set.
QM9 slices use $|\text{logP}-\text{logP}_{\text{cond}}|\leq 1$;
atom counts include hydrogens.
Lowest errors and means closest to their target or reference are
bolded using unrounded values. $W_1$ at target $+4$ is omitted.}
\label{tab:logp-baselines}
\end{minipage}\hfill
\begin{minipage}[c]{0.60\linewidth}
\centering
\footnotesize
\setlength{\tabcolsep}{2.4pt}
\begin{tabular}{@{}llccccc@{}}
\toprule
Metric & Model & \multicolumn{5}{c}{$\text{logP}_{\text{cond}}$} \\
\cmidrule(lr){3-7}
& & $-4$ & $-2$ & $0$ & $2$ & $4$ \\
\midrule
Mean $n_{\text{atoms}}$ & QM9 slice & 16.5 & 15.9 & 17.1 & 20.0 & 26.1 \\
Mean logP & QM9 slice & $-3.62$ & $-1.39$ & 0.07 & 1.57 & 3.16 \\
\midrule
\multirow{3}{*}{logP MAE $\downarrow$}
& \morph & 1.20 & 0.46 & 0.30 & 0.31 & \textbf{0.27} \\
& SemlaFlow, $p(n)$ & 0.98 & 0.48 & 0.29 & 0.28 & 1.15 \\
& SemlaFlow, oracle & \textbf{0.82} & \textbf{0.41} & \textbf{0.26} & \textbf{0.21} & 0.82 \\
\midrule
\multirow{3}{*}{Mean logP}
& \morph & $-2.91$ & $-1.66$ & 0.08 & \textbf{2.04} & \textbf{3.74} \\
& SemlaFlow, $p(n)$ & $-3.13$ & $-1.78$ & 0.05 & 1.84 & 2.85 \\
& SemlaFlow, oracle & $\mathbf{-3.29}$ & $\mathbf{-1.84}$ & \textbf{0.02} & 1.93 & 3.18 \\
\midrule
\multirow{3}{*}{Mean $n_{\text{atoms}}$}
& \morph & \textbf{16.8} & \textbf{16.4} & 17.1 & 21.2 & 31.8 \\
& SemlaFlow, $p(n)$ & 18.0 & 18.0 & 18.0 & 18.1 & 18.0 \\
& SemlaFlow, oracle & 17.0 & 17.3 & \textbf{17.1} & \textbf{21.0} & \textbf{25.3} \\
\midrule
\multirow{3}{*}{$W_1(n_{\text{atoms}})$ $\downarrow$}
& \morph & 0.6 & \textbf{0.5} & 0.4 & 1.2 & -- \\
& SemlaFlow, $p(n)$ & 1.6 & 2.1 & 0.9 & 1.9 & -- \\
& SemlaFlow, oracle & \textbf{0.5} & 1.4 & \textbf{0.3} & \textbf{0.9} & -- \\
\bottomrule
\end{tabular}
\end{minipage}
\end{table}

\begingroup
\setlength{\intextsep}{6pt}
\subsection{Multi-property conditioning QED-logP}
\label{sec:apx_logp_qed}
We perform additional experiments on multi-property conditioning to evaluate
\morph's ability to generate molecules satisfying joint property requirements.
We combine QED, which measures drug-likeness, with logP, using the target grid
$\mathrm{logP}_{\mathrm{cond}}\in\{-2,0,2\}$ and
$\mathrm{QED}_{\mathrm{cond}}\in\{0.3,0.5,0.7\}$, with 10,000 generation
attempts per pair. Figure~\ref{fig:logp_qed} and
Table~\ref{tab:cfg_merged_qed_logp_compact}
report the results. The $(-2,0.7)$ setting yields 11.74\% validity and 5.28\%
PoseBusters success, but no output meets both targeting tolerances.

\begin{figure}[H]
    \centering
    \includegraphics[width=0.98\linewidth]{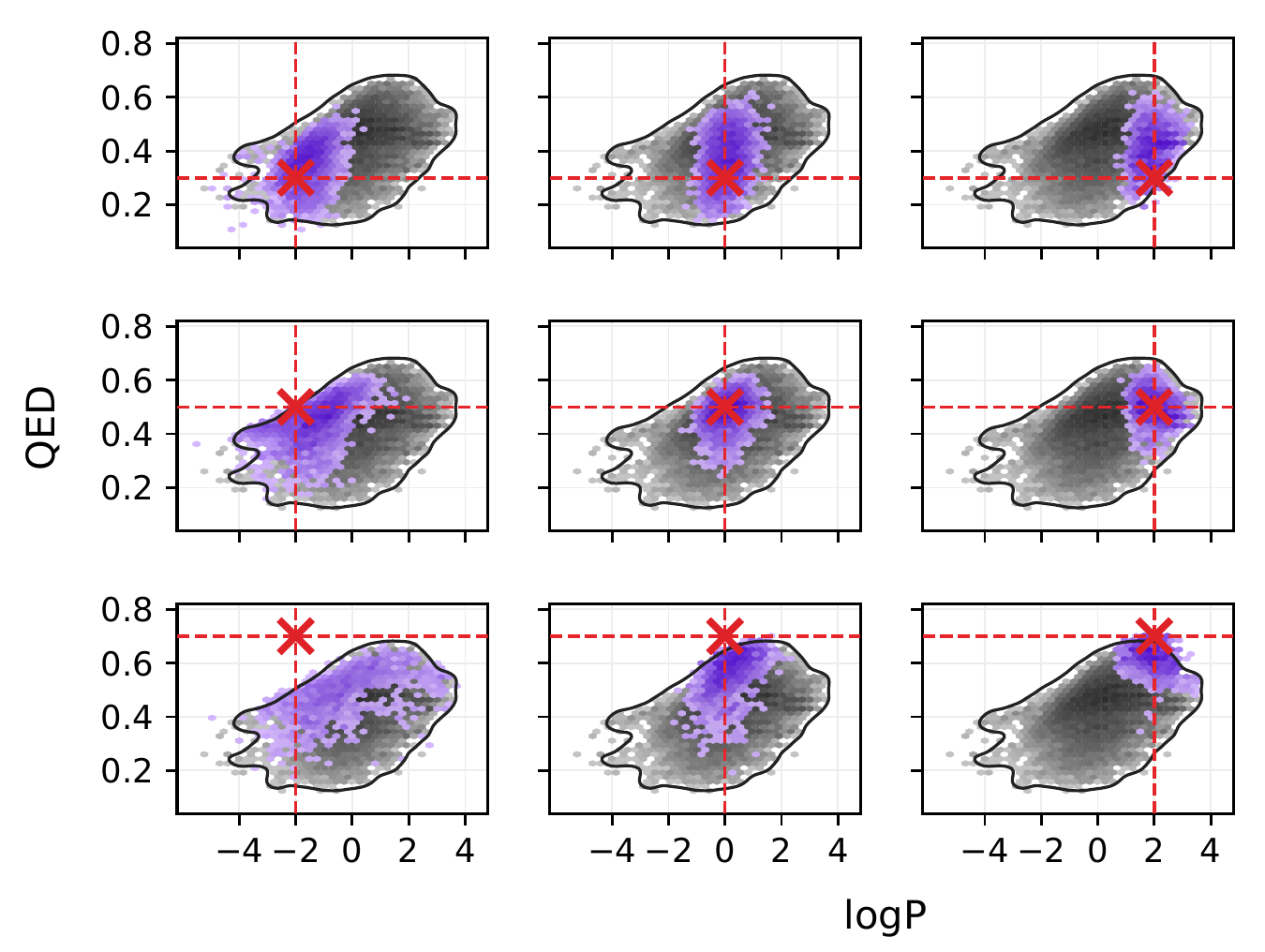}
    \caption{Joint QED--logP conditioning on QM9, with 10,000 generation
    attempts per target pair. Generated property distributions are shown in
    purple against the training distribution in gray. Red crosses and dashed
    lines mark the targets: logP values $-2,0,2$ across columns and QED values
    $0.3,0.5,0.7$ down rows.}
    \label{fig:logp_qed}
\end{figure}

\begin{table}[H]
\centering
\small
\setlength{\tabcolsep}{2pt}
\captionsetup{font=small}
\caption{Joint QED--logP conditioning. Rates are percentages.
Valid is connected sanitization; PB requires all 12 PoseBusters checks.
Scored denotes outputs with finite logP and QED values.
Valid, PB, Scored, and Joint use all attempts. Uniqueness and novelty use
SMILES-convertible outputs; stability uses the valence table.
Means ($\mu$), SDs, and MAEs use scored outputs within each run.
Joint requires both $|\mathrm{logP}-\mathrm{logP}_{\mathrm{cond}}|\leq 0.5$
and $|\mathrm{QED}-\mathrm{QED}_{\mathrm{cond}}|\leq 0.05$.}
\begin{tabular}{@{}cc|rrrrrr|r|cr|cr|r@{}}
\toprule
\multicolumn{2}{c|}{Targets} & \multicolumn{6}{c|}{Generative quality (\%)} & Scored & \multicolumn{2}{c|}{logP} & \multicolumn{2}{c|}{QED} & Joint \\
logP & QED & Valid & PB & Uniq. & Nov. & \shortstack{Atom\\stab.} & \shortstack{Mol.\\stab.} & (\%) & $\mu$ & MAE & $\mu$ & MAE & (\%) \\
\midrule
-2.0 & 0.30 & 96.99 & 89.71 & 63.94 & 46.14 & 99.97 & 96.57 & 97.05 & -1.66$_{\pm0.46}$ & 0.453 & 0.36$_{\pm0.06}$ & 0.070 & 28.04 \\
-2.0 & 0.50 & 87.82 & 81.75 & 66.73 & 74.58 & 99.80 & 85.09 & 88.26 & -1.32$_{\pm0.76}$ & 0.869 & 0.45$_{\pm0.06}$ & 0.059 & 8.91 \\
-2.0 & 0.70 & 11.74 & 5.28 & 94.36 & 99.20 & 96.16 & 7.87 & 17.56 & -0.42$_{\pm1.35}$ & 1.684 & 0.49$_{\pm0.08}$ & 0.210 & 0.00 \\
\midrule
0.0 & 0.30 & 98.85 & 91.87 & 77.86 & 22.06 & 99.99 & 98.77 & 98.90 & 0.07$_{\pm0.38}$ & 0.307 & 0.37$_{\pm0.07}$ & 0.082 & 29.15 \\
0.0 & 0.50 & 98.95 & 94.43 & 84.74 & 24.51 & 99.99 & 98.86 & 98.99 & 0.08$_{\pm0.39}$ & 0.324 & 0.49$_{\pm0.04}$ & 0.030 & 66.35 \\
0.0 & 0.70 & 88.72 & 87.28 & 39.18 & 68.36 & 99.94 & 88.68 & 89.63 & 0.23$_{\pm0.46}$ & 0.408 & 0.59$_{\pm0.05}$ & 0.109 & 2.03 \\
\midrule
2.0 & 0.30 & 97.02 & 93.28 & 46.41 & 18.85 & 99.99 & 96.91 & 97.12 & 1.98$_{\pm0.38}$ & 0.304 & 0.40$_{\pm0.06}$ & 0.099 & 19.81 \\
2.0 & 0.50 & 96.39 & 92.71 & 55.76 & 11.59 & 99.99 & 96.28 & 96.54 & 2.04$_{\pm0.41}$ & 0.336 & 0.50$_{\pm0.04}$ & 0.030 & 63.61 \\
2.0 & 0.70 & 88.72 & 87.89 & 25.99 & 44.22 & 99.93 & 88.81 & 89.87 & 1.94$_{\pm0.39}$ & 0.310 & 0.64$_{\pm0.04}$ & 0.064 & 31.67 \\
\bottomrule
\end{tabular}
\label{tab:cfg_merged_qed_logp_compact}
\end{table}
\endgroup
\clearpage

\section{Interpolation examples}
\label{sec:interpolation_mols}
\begin{figure}[H]
    \centering
    \includegraphics[width=0.9\linewidth]{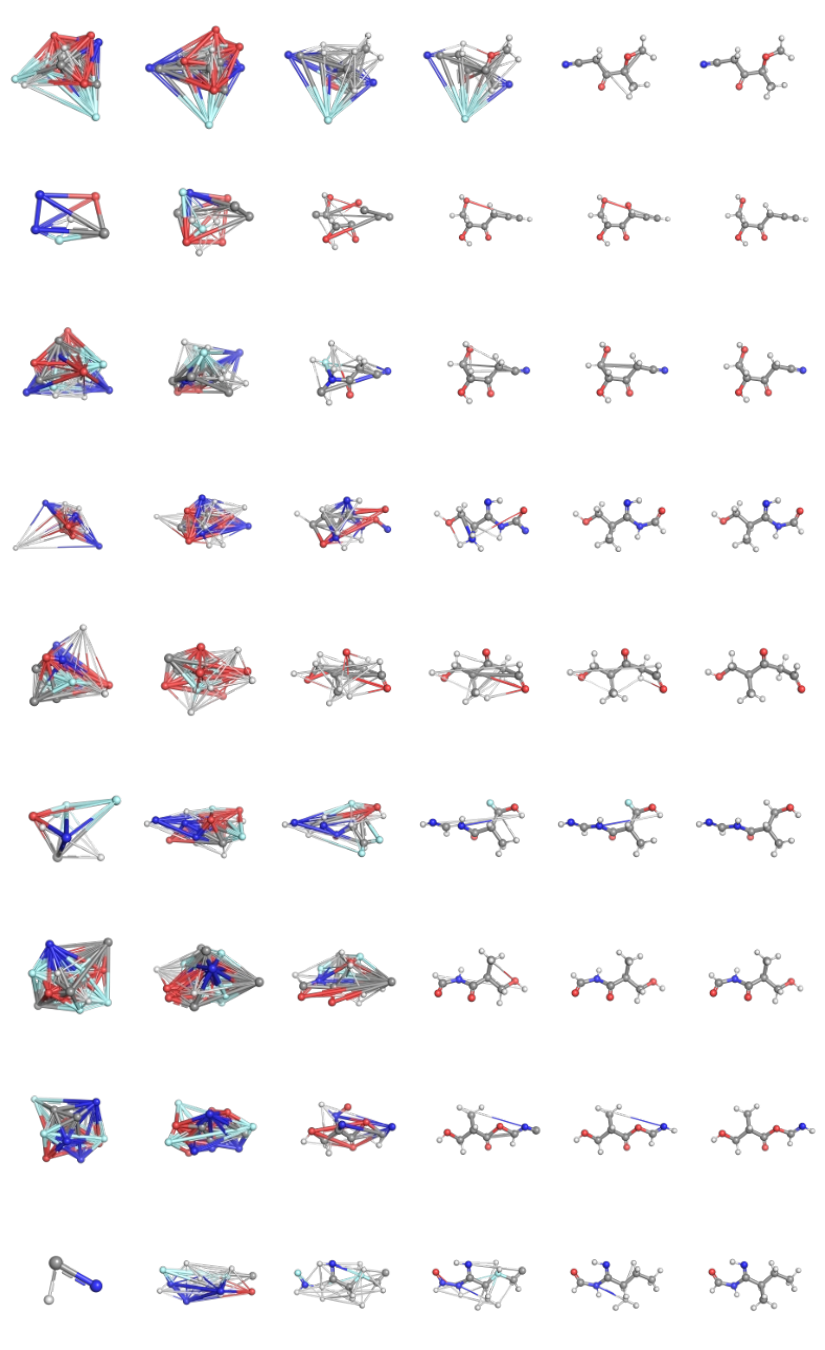}
    \caption{Randomly selected QM9 interpolation trajectories. }
    \label{fig:interpolation_mols}
\end{figure}

\section{Sampled Molecules}
\label{sec:sampled_mols}
\begin{figure}[H]
    \centering
    \includegraphics[width=0.9\linewidth]{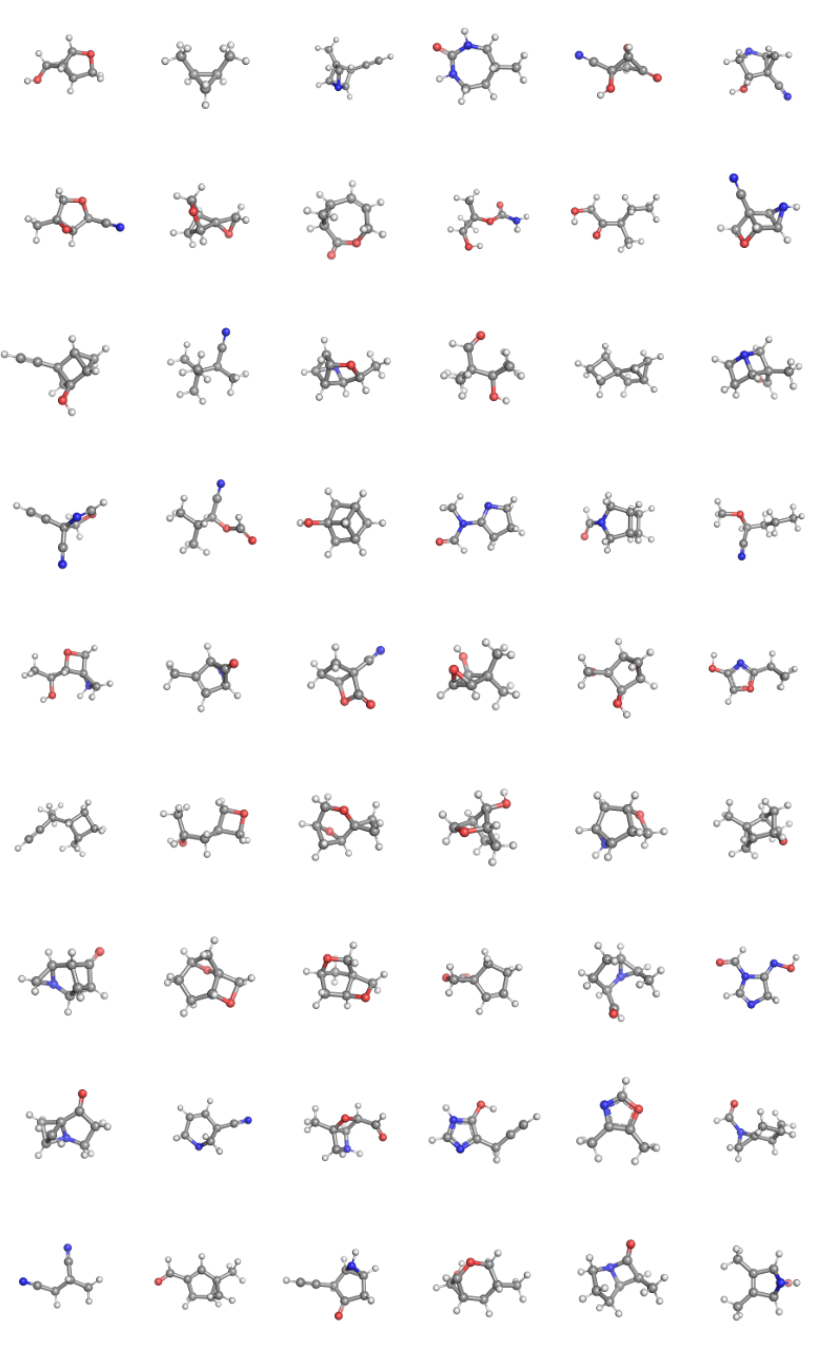}
    \caption{Sampled valid molecules using \morph trained on QM9.}
    \label{fig:qm9-unconditional-samples}
\end{figure}

\begin{figure}[H]
    \centering
    \includegraphics[width=0.9\linewidth]{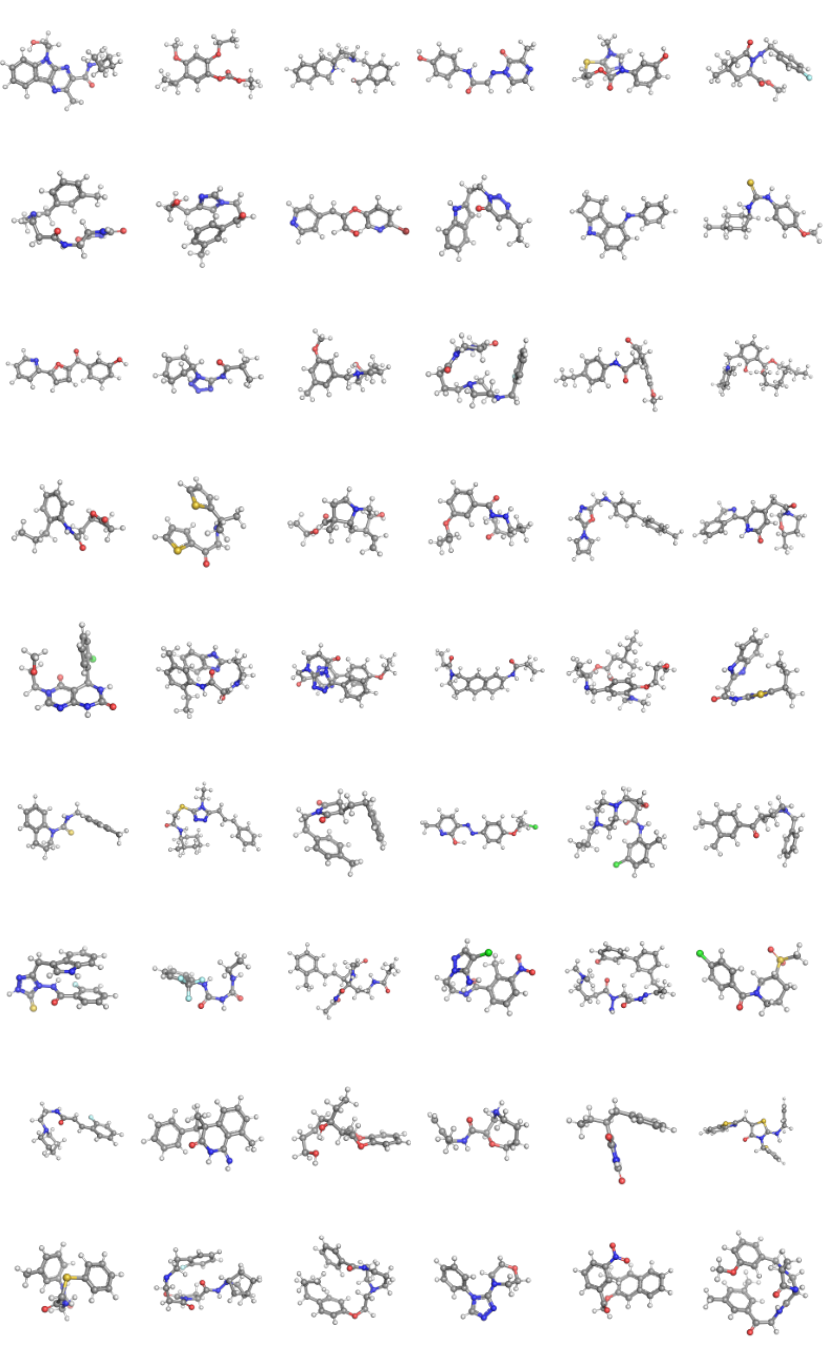}
    \caption{Sampled valid molecules using \morph trained on GEOM-Drugs.}
    \label{fig:geom-unconditional-samples}
\end{figure}

\clearpage



\end{document}